\documentclass[runningheads,a4paper]{llncs}

\usepackage{times}
\usepackage{epsfig}
\usepackage{graphicx}
\usepackage{amsmath, amssymb}
\usepackage{algorithm, algorithmic, url, enumerate}
\usepackage{color}
\usepackage[width=122mm,left=12mm,paperwidth=146mm,height=193mm,top=12mm,paperheight=217mm]{geometry}

\usepackage{multirow}
\usepackage{colortbl}
\definecolor{orangecolor1}{rgb}{0.9686,0.5882,0.2745}
\definecolor{orangecolor2}{rgb}{0.9882,0.8667,0.8118}
\definecolor{orangecolor3}{rgb}{0.9922,0.9373,0.9137}

\newcommand{\affm}{\mathbf{W}} 

\newcommand{\cluster}{\mathcal{C}}

\newtheorem{thm}{Theorem} 
\newtheorem{lem}[thm]{Lemma}

\newtheorem{defn}{Definition} 
\newtheorem{rem}{Remark}

\urldef{\mailsa}\path|{alfred.hofmann, ursula.barth, ingrid.haas, frank.holzwarth,|
\urldef{\mailsb}\path|anna.kramer, leonie.kunz, christine.reiss, nicole.sator,|
\urldef{\mailsc}\path|erika.siebert-cole, peter.strasser, lncs}@springer.com|

\begin{document}

\mainmatter  

\title{Graph Degree Linkage: \\Agglomerative Clustering on a Directed Graph}

\titlerunning{Graph Degree Linkage: Agglomerative Clustering on a Directed Graph}

\author{Wei Zhang\inst{1} \and %
Xiaogang Wang\inst{2,3} \and %
Deli Zhao\inst{1} \and %
Xiaoou Tang\inst{1,3}}
\institute{  
  Department of Information Engineering, %
  The Chinese University of Hong Kong %
  \email{wzhang009@gmail.com} %
  \and
  Department of Electronic Engineering, %
  The Chinese University of Hong Kong %
  \and
  Shenzhen Institutes of Advanced Technology, %
  Chinese Academy of Sciences, China %
}

\authorrunning{W. Zhang, X. Wang, D. Zhao and X. Tang}

\maketitle

\setcounter{footnote}{0}

\begin{abstract}
This paper proposes a simple but effective graph-based agglomerative algorithm, for clustering high-dimensional data. We explore the different roles of two fundamental concepts in graph theory, indegree and outdegree, in the context of clustering. The average indegree reflects the density near a sample, and the average outdegree characterizes the local geometry around a sample. Based on such insights, we define the affinity measure of clusters via the product of average indegree and average outdegree. The product-based affinity makes our algorithm robust to noise. The algorithm has three main advantages: good performance, easy implementation, and high computational efficiency. We test the algorithm on two fundamental computer vision problems: image clustering and object matching. Extensive experiments demonstrate that it outperforms the state-of-the-arts in both applications.\footnote{The code and supplemental materials are publicly available at \url{http://mmlab.ie.cuhk.edu.hk/research/gdl/}.}
\end{abstract}

\begin{figure}[t]
\center
{\footnotesize
\begin{tabular}{l}
\includegraphics[width=0.766\columnwidth]{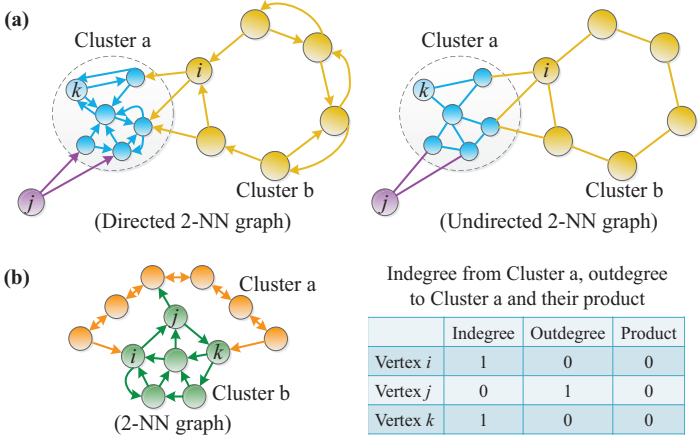} 
\end{tabular}
}
\caption{(a) Indegree can be use to detect the change of densities. The density in Cluster a is high, and the density in Cluster b is low. The vertices inside Cluster a are strongly connected, but there is no outedge to vertices outside Cluster a. So the indegree of $k$ from Cluster a is nonzero, while the indegree of $i$ (a vertex in Cluster b) and $j$ (an outlier) from Cluster a are zero. If an undirected graph is considered without separating indegrees and outdegrees, both $i$ and $j$ have the same degree from Cluster a as $k$. (b) The product of the indegree and outdegree is an affinity measure robust to noisy edges between the two clusters. Under this measure, Cluster a and Cluster b have a zero affinity, i.e., the sum of product of indegree and outdegree for all vertices is 0, and thus they are separated well.} \label{fig:toygraph}
\end{figure}

\section{Introduction} \label{sec:intro}

Many problems in computer vision involve clustering. Partitional clustering, such as $k$-means \cite{hastie2009elements}, determines all clusters at once, while agglomerative clustering \cite{hastie2009elements} begins with a large number of small clusters, and iteratively selects two clusters with the largest affinity under some measures to merge, until some stopping condition is reached. Agglomerative clustering has been studied for more than half a century, and used in many applications \cite{hastie2009elements}, because it is conceptually simple and produces an informative hierarchical structure of clusters.



Classical agglomerative clustering algorithms have several limitations \cite{hastie2009elements}, which have restricted their wider applications in computer vision. The data in computer vision applications are usually high dimensional. The distributions of data clusters are often in different densities, sizes, and shapes, and form manifold structures. In addition, there are often noise and outliers in data. The conventional agglomerative clustering algorithms, such as the well-known linkage methods \cite{hastie2009elements}, usually fail to tackle these challenges. As their affinities are directly computed using pairwise distances between samples and cannot capture the global manifold structures in high-dimensional spaces, these algorithms have problems of clustering high-dimensional data, and are quite sensitive to noise and outliers \cite{hastie2009elements}. 

To tackle these problems, we propose a simple and fast graph-based agglomerative clustering algorithm. The graph representation of data has been extensively exploited in various machine learning topics \cite{shi2000ncut,ng2001spectral,belkin2003laplacian,grady2006isoperimetric,zhang2011srmm}, but has rarely been utilized in agglomerative clustering. Our algorithm builds $K$-nearest-neighbor ($K$-NN) graphs using the pairwise distances between samples, since studies \cite{belkin2003laplacian} show the effectiveness of using local neighborhood graphs to model data lying on a low-dimensional manifold embedded in a high-dimensional space.

\begin{figure}[t]
\center
{\footnotesize
\begin{tabular}{cccc}
\includegraphics[width=0.136\columnwidth]{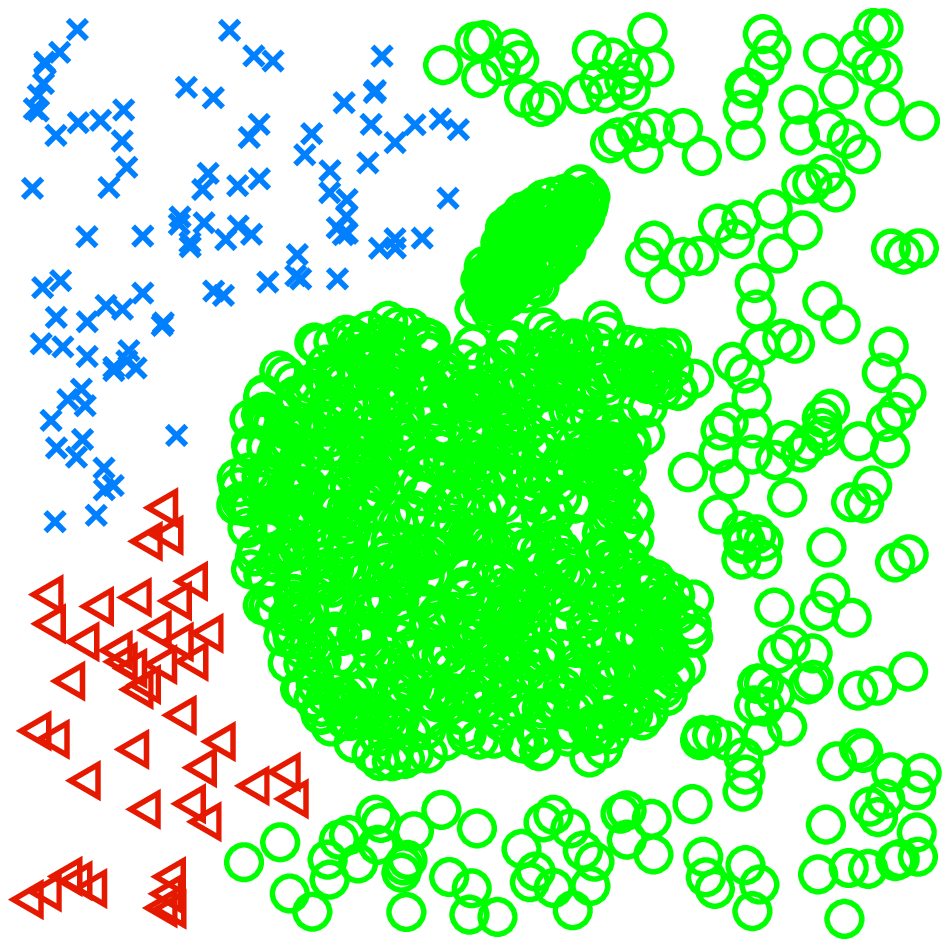}
& \includegraphics[width=0.136\columnwidth]{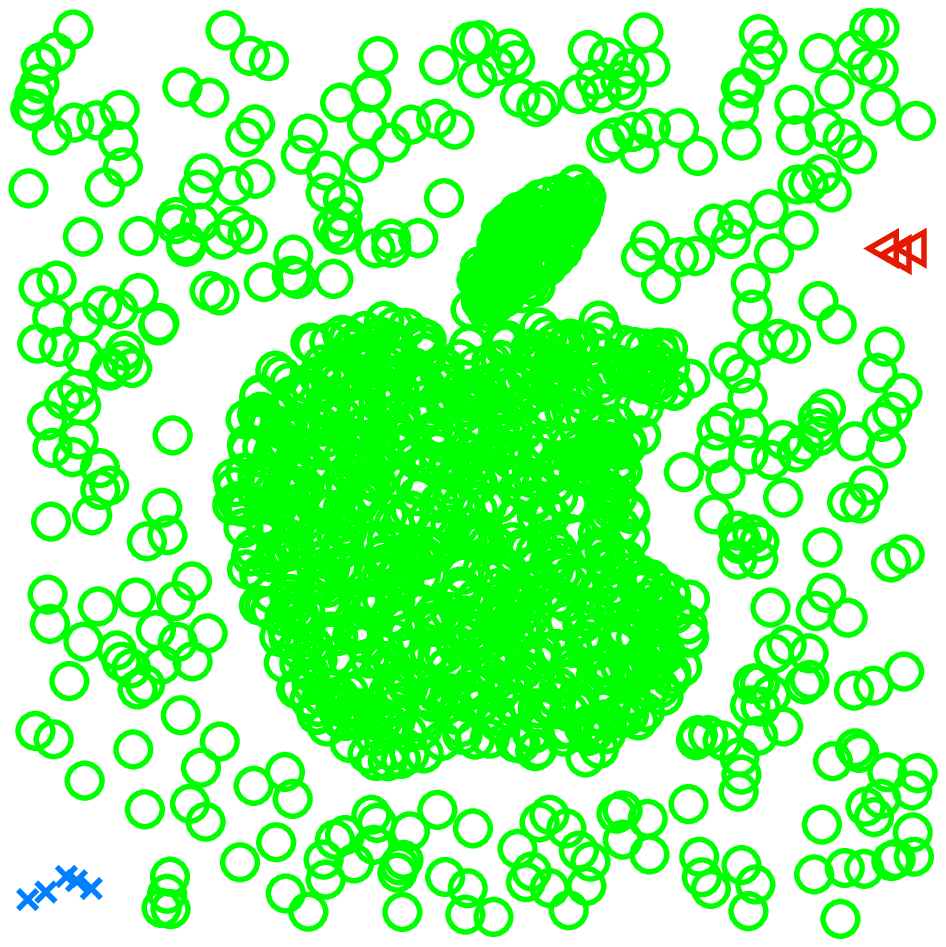}
& \includegraphics[width=0.136\columnwidth]{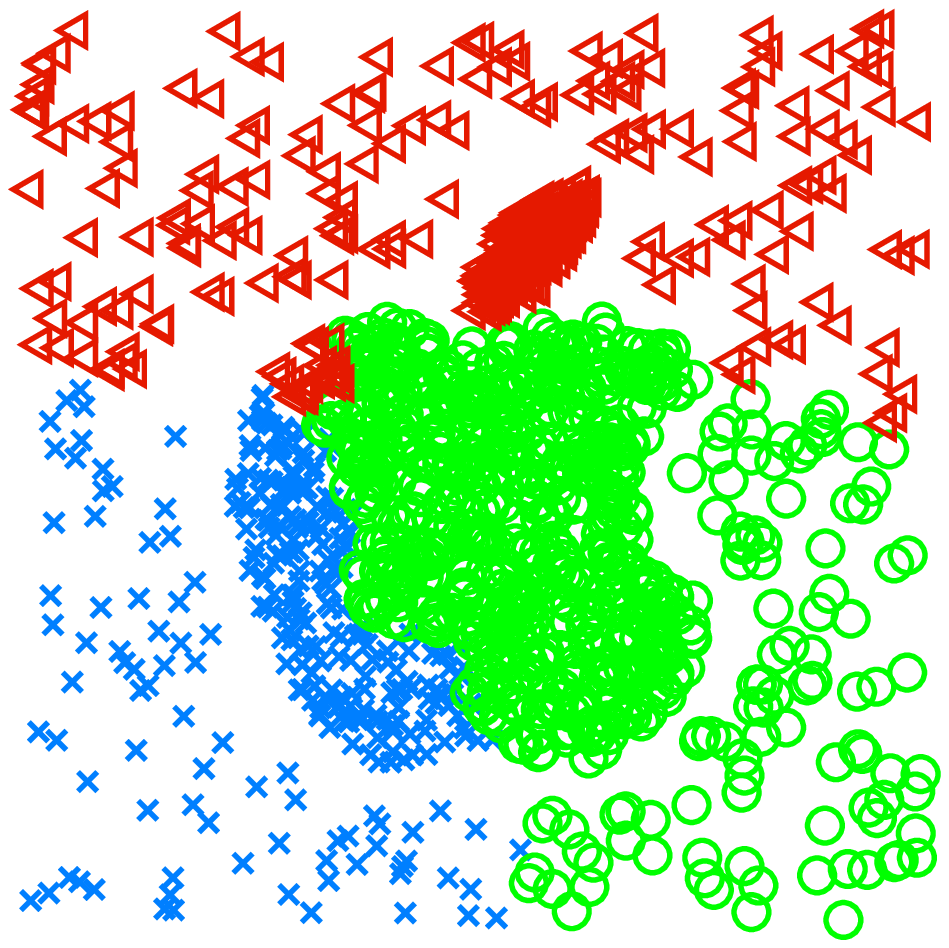}
& \includegraphics[width=0.136\columnwidth]{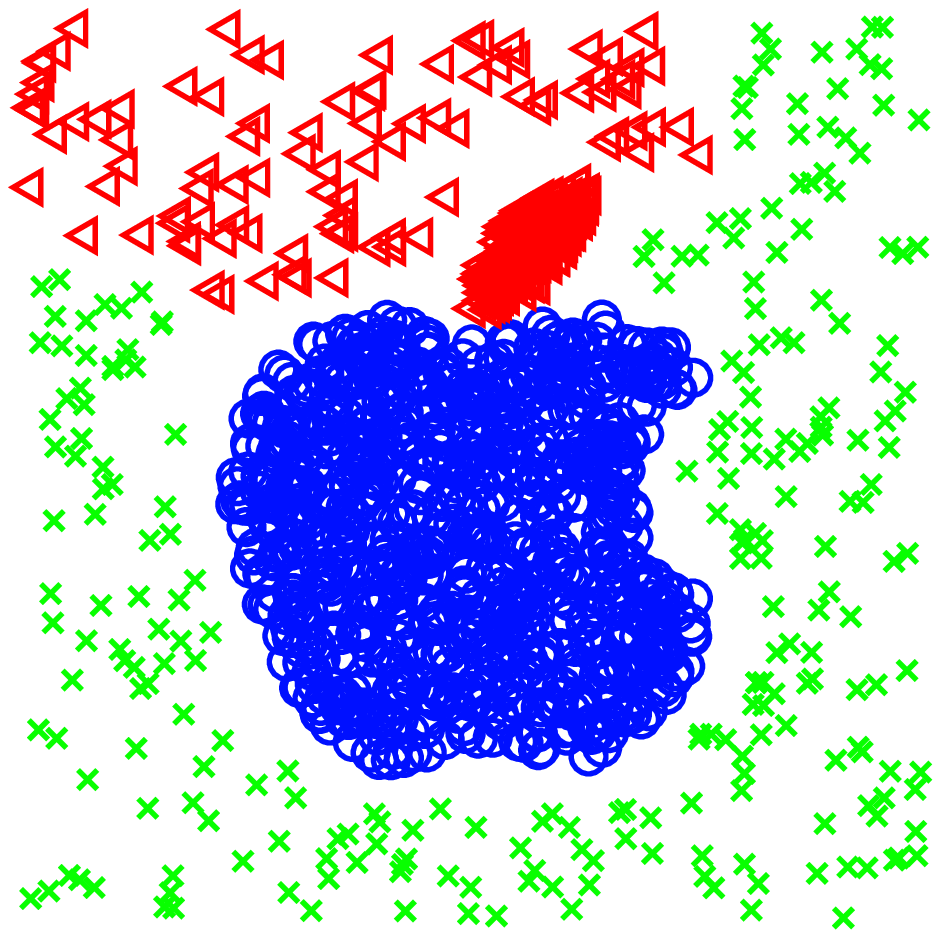} \\
\makebox[0.236\columnwidth][c]{Average Linkage}
& \makebox[0.236\columnwidth][c]{Single Linkage}
& \makebox[0.236\columnwidth][c]{Complete Linkage}
& \makebox[0.236\columnwidth][c]{Graph-based Linkage} \vspace{5pt} \\
\includegraphics[width=0.136\columnwidth]{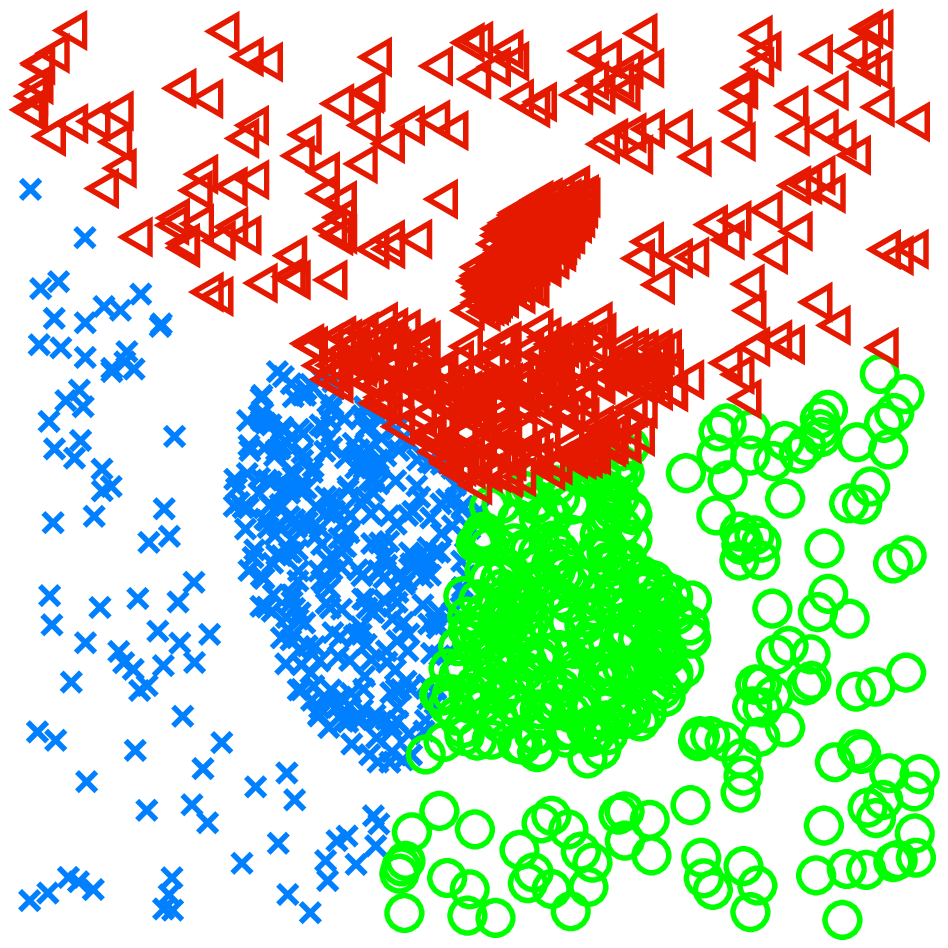}
& \includegraphics[width=0.136\columnwidth]{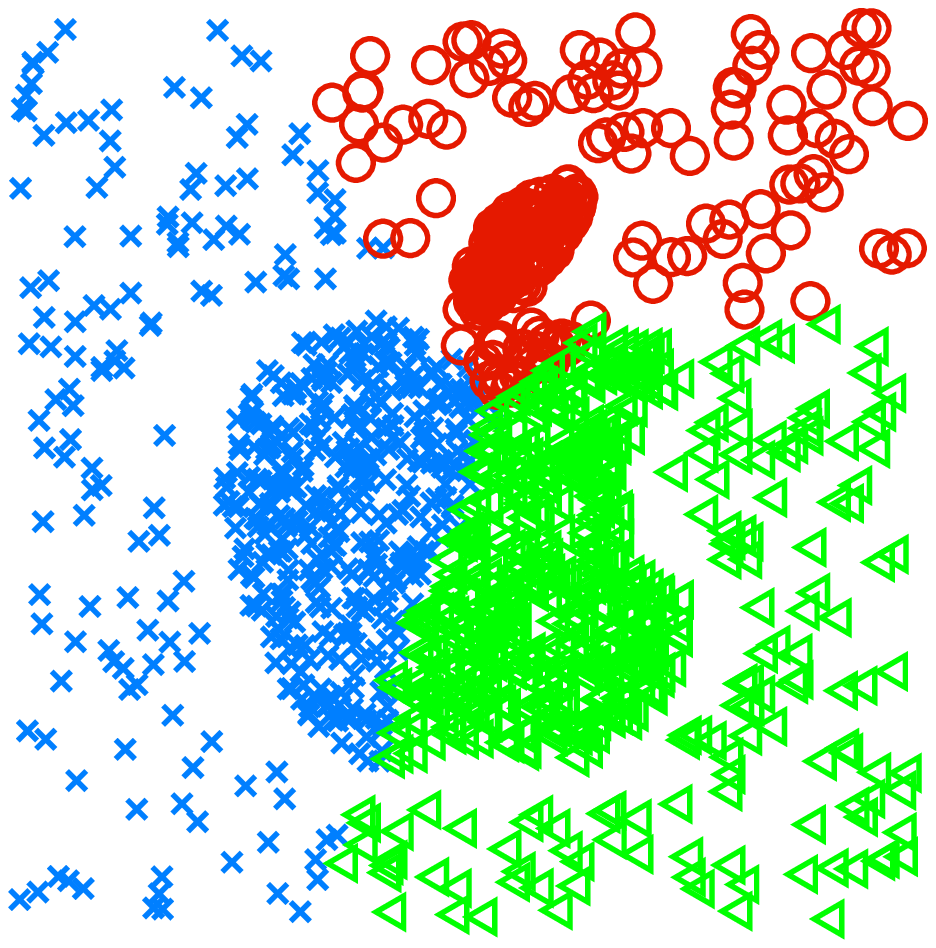}
& \includegraphics[width=0.136\columnwidth]{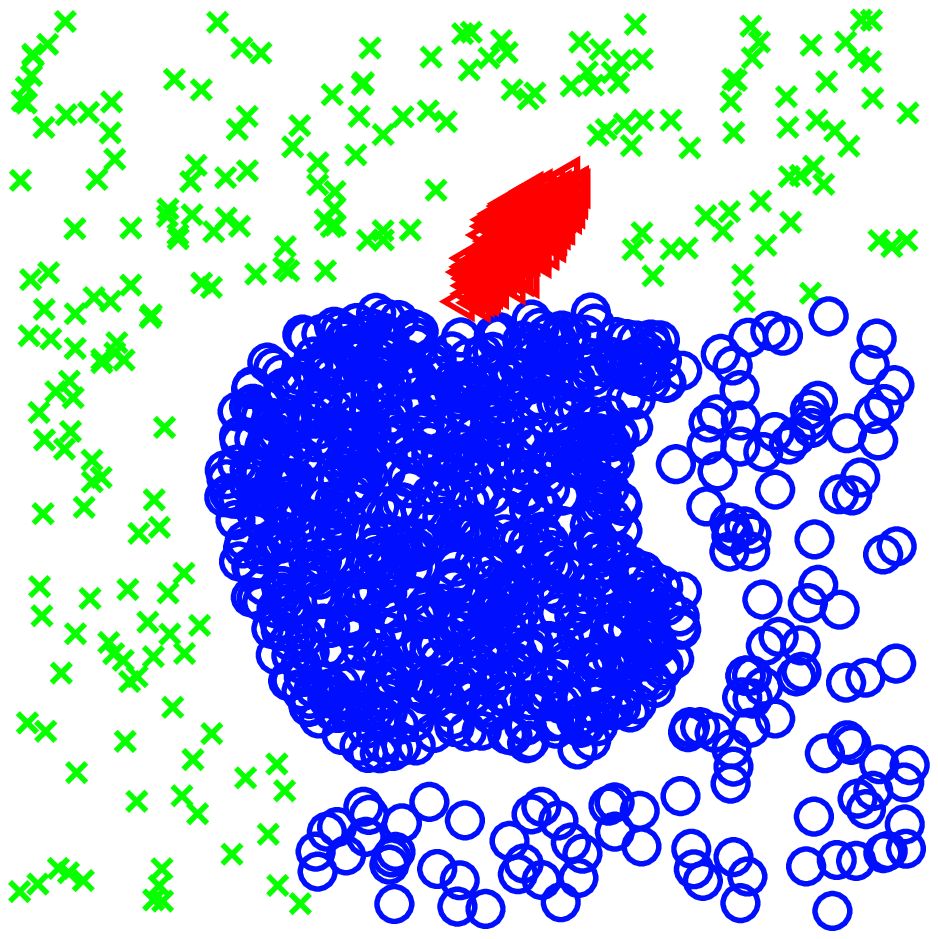}
& \includegraphics[width=0.136\columnwidth]{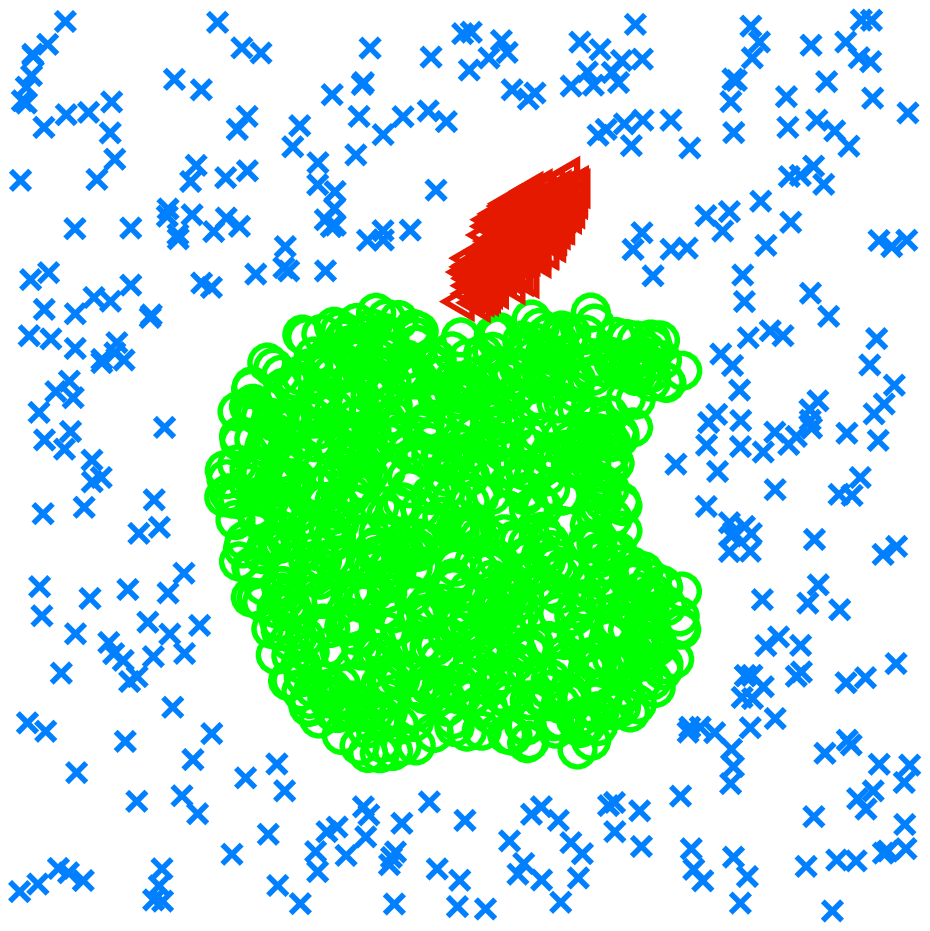} \\
\makebox[0.236\columnwidth][c]{AP \cite{frey2007ap}}
& \makebox[0.236\columnwidth][c]{SC \cite{ng2001spectral}}
& \makebox[0.236\columnwidth][c]{DGSC \cite{zhou2005directed}}
& \makebox[0.236\columnwidth][c]{Ours} \\
\end{tabular}
}
\caption{Results of different clustering algorithms on a synthetic multiscale dataset. Our algorithm can perfectly discover the three clusters with different shapes, sizes, and densities. The output clusters are shown in color (best viewed on screen).} \label{fig:toyexp}
\end{figure}

We use the \emph{indegree} and \emph{outdegree}, fundamental concepts in graph theory, to characterize the affinity between two clusters. The outdegree of a vertex to a cluster measures the similarity between the vertex and the cluster. If many of the $K$-NNs of the vertex belong the cluster, the outdegree is large. The outdegree can capture the manifold structures in the high dimensional space. The indegree of a vertex from a cluster reflects the density near the vertex. It is effective for detecting the change of densities, which often occurs at the boundary of clusters. Therefore, we use it to separate clusters close in space but different in densities, and also reduce the effect of noise. An example is shown in Fig. \ref{fig:toygraph}(a). To our best knowledge, properties of the indegree and outdegree have not been explored by any existing clustering algorithm, although they were successfully applied in analysis of complex networks such as World Wide Web \cite{kleinberg1999authoritative} and social networks \cite{mislove2007measurement} and showed interesting results.



Our affinity measure between two clusters is defined as follows. First, the structural affinity from a vertex to a cluster is defined via the product of the average indegree from the cluster and average outdegree to the cluster. Intuitively, if a vertex belongs to a cluster, it should be strongly connected to the cluster, i.e., both its indegree and outdegree are large. Otherwise, either the indegree or outdegree is small. Therefore, the product of indegree and outdegree can be a good affinity measure (Fig. \ref{fig:toygraph}(b)). We show that the correlation between the inter-cluster indegree and outdegree is weak across different vertices, if the two clusters belong to different ground-truth clusters, using synthetic data in Fig. \ref{fig:product}. Then, the affinity between two clusters is naturally the aggregated affinity measure for all the vertices in the two clusters.


Our algorithm has three main advantages as follows.

First of all, it has \emph{outstanding performance}, especially on noisy data and multiscale data (i.e., clusters in different densities). The visual comparisons with linkage methods \cite{hastie2009elements}, graph-based average linkage, affinity propagation (AP) \cite{frey2007ap}, spectral clustering (SC) \cite{ng2001spectral}, and directed graph spectral clustering (DGSC) \cite{zhou2005directed} on synthetic multiscale data are shown in Fig. \ref{fig:toyexp}. Noise and multiple scales can degrade the performance of spectral clustering greatly \cite{zelnik2004self}, while the indegree and outdegree in our algorithm detect the boundary of scales automatically\footnote{E.g., if cluster $a$ has higher density than cluster $b$, the boundary of cluster $a$ will have high indegree and low outdegree, while the boundary of cluster $b$ will have low indegree and high outdegree.} and reduce the effect of noise. In Sec. \ref{sec:exp}, extensive experiments on real data, including imagery data and feature correspondence data, demonstrate its superiority over state-of-the-art methods. These experiments aim at two fundamental problems in computer vision, i.e., image clustering and object matching, and the results suggest many potential applications of our work.

Second, it is \emph{easy to implement}. This affinity measure can be expressed in a matrix form and implemented with vector additions and inner-products. Therefore, our algorithm can be implemented without any dependency on external numerical libraries, such as eigen-decomposition which was extensively employed by many clustering algorithms \cite{shi2000ncut,ng2001spectral,wu2007local}.

Finally, it is \emph{very fast}. We propose an acceleration method for our algorithm. In practice, our algorithm is much faster than spectral clustering \cite{shi2000ncut,ng2001spectral}, especially on large-scale data.



\section{Related Work}
The literature dedicated to agglomerative clustering is abundant \cite{hastie2009elements,franti2006fast,cho2009feature}. Linkages \cite{hastie2009elements}, e.g., average linkage, define the affinity based on pairwise distances between samples. Since pairwise distances do not well capture the global structures of data, these methods fail on clustering data with complex structures and are sensitive to noise \cite{hastie2009elements} (see the example in Fig. \ref{fig:toyexp}). Many variants of linkage methods, such as DBSCAN \cite{sander1998density}, have been proposed in the data mining community and show satisfactory performance. However, they usually fail to tackle the great challenge from high-dimensional spaces, because their sophisticated affinity measures are based on observations from low-dimensional data \cite{ertoz2003finding}.


Several algorithms \cite{karypis1999chameleon,zhao2008zeta,felzenszwalb2004efficient} has attempted to perform agglomerative clustering on the graph representation of data. Chameleon \cite{karypis1999chameleon} defines the cluster affinity from relative interconnectivity and relative closeness, both of which are based on a min-cut bisection of clusters. Although good performance was shown on 2D toy datasets, it suffers from high computational cost because its affinity measure is based on a min-cut algorithm. Zell \cite{zhao2008zeta} describes the structure of a cluster via the zeta function and defined the affinity based on the structural changes after merging. It needs to compute matrix inverse in each affinity computation, so it is much slower than our simple algorithm (see Sec. \ref{sec:exp_img}). Felzenszwalb and Huttenlocher proposed an effective algorithm for image segmentation \cite{felzenszwalb2004efficient}.

Besides agglomerative clustering, $K$-means \cite{hastie2009elements} and spectral clustering \cite{shi2000ncut,ng2001spectral,yu2003multiclass} are among the most widely used clustering algorithms. However, $K$-means is sensitive to the initialization and difficult to handle clusters with varying densities and sizes, or manifold shapes. Although spectral clustering can handle the manifold data well, its performance usually degrades greatly with the existence of noise and outliers, because the eigenvectors of graph Laplacian are sensitive to noisy perturbations \cite{grady2006isoperimetric}. Affinity Propagation \cite{frey2007ap} explores the intrinsic data structures by message passing among data points. Although it performs well on high-dimensional data, it usually requires considerable run-time, especially when the preference value cannot be manually set.

Directed graphs have been studied for spectral clustering (e.g., \cite{zhou2005directed}). However, these methods symmetrize the directed graph before the clustering task. In contrast, we only symmetrize the affinity between two clusters, while keep the directed graph during the clustering process. Therefore, our algorithm utilizes more information from the asymmetry and is more robust to noisy edges (see Fig. \ref{fig:toyexp} for a comparison between DGSC \cite{zhou2005directed} and our algorithm).

\section{Graph Degree Linkage} \label{sec:gdl}


\subsection{Neighborhood Graph} \label{sec:nngraph}
Given a set of samples $\mathcal{X} = \{\mathbf{x}_1 , \mathbf{x}_2 , ... , \mathbf{x}_n\}$, we build a directed graph $G = (V, E)$, where $V$ is the set of vertices corresponding to the samples in $\mathcal{X}$ , and $E$ is the set of edges connecting vertices. The graph is associated with a weighted adjacency matrix $\mathbf{W} = [w_{ij}]$, where $w_{ij}$ is the weight of the edge from vertex $i$ to vertex $j$. $w_{ij} = 0$ if and only if there is no edge from $i$ to $j$.

To capture the manifold structures in high-dimensional spaces, we use the $K$-NN graph, in which the weights are defined as
\begin{equation}  \label{eqn:graphw}
w_{ij} = \left\{\begin{array}{ll}
\exp\left(- \frac{dist(i,j)^2}{\sigma^2} \right), & \mbox{if } \mathbf{x}_j \in \mathcal{N}_i^K, \\
 0, & \mbox{otherwise},
\end{array} \right.
\end{equation}
where $dist(i,j)$ is the distance between $\mathbf{x}_i$ and $\mathbf{x}_j$, $\mathcal{N}_i^K$ is the set of $K$-nearest neighbors of $\mathbf{x}_i$, and $\sigma^2$ is set as $\sigma^2 = \frac{a}{nK} \left[\sum_{i=1}^n \sum_{\mathbf{x}_j \in \mathcal{N}_i^K} dist(i, j)^2 \right]$. $K$ and $a$ are free parameters to be set. In a $K$-NN graph, there is an edge pointing from $\mathbf{x}_i$ to $\mathbf{x}_j$ with weight $w_{ij}$, if $\mathbf{x}_j \in \mathcal{N}_i^K$.




\subsection{Algorithm Overview} \label{sec:overview}
The graph degree linkage (GDL) algorithm begins with a number of initial small clusters, and iteratively selects two clusters with the maximum affinity to merge. The affinities are computed on the $K$-NN graph, based on the indegree and outdegree of vertices in the two clusters.

The initial small clusters are simply constructed as \emph{weakly connected components} of a $K^0$-NN graph, where the neighborhood size $K^0$ is small, typically as $1$ or $2$. Then, each component is an initial cluster, and each sample is assigned to only one cluster.
\begin{defn}
A connected component of an undirected graph is a maximal connected subgraph in which any two vertices are connected to each other by paths.

A weakly connected component of a directed graph is a connected component of the undirected graph produced by replacing all of its directed edges with undirected edges.
\end{defn}

The GDL algorithm is presented as Algorithm \ref{alg:gdl}, with details given in the following subsection.

\begin{algorithm}
\caption{Graph Degree Linkage (GDL)} \label{alg:gdl}
\begin{algorithmic}
\STATE \textbf{Input:} a set of $n$ samples $\mathcal{X} = \{\mathbf{x}_1, \mathbf{x}_2, \cdots, \mathbf{x}_n\}$, and the target number of clusters $n_T$.
\STATE Build the $K^0$-NN graph, and detect its weakly connected components as initial clusters. Denote the set of initial clusters as $V^c = \{\cluster_1, \cdots, \cluster_{n_c}\}$, where $n_c$ is the number of clusters.
\STATE Build the $K$-NN graph, and get the weighted adjacency matrix $\mathbf{W}$.
\WHILE{$n_c > n_T$}
\STATE Search two clusters $\cluster_a$ and $\cluster_b$, such that $\{\cluster_a,\cluster_b\} = \operatornamewithlimits{argmax}_{\cluster_a, \cluster_b \in V^c} \mathcal{A}_{\cluster_a, \cluster_b}$, where $\mathcal{A}_{\cluster_a, \cluster_b}$ is the affinity measure between $\cluster_a$ and $\cluster_b$, computed using Eq. (\ref{eqn:aff}).
\STATE $V^c \leftarrow \{V^c \setminus \{\cluster_a,\cluster_b\}\} \cup \{\cluster_a \cup \cluster_b\}$, and $n_c = n_c - 1$.
\ENDWHILE
\STATE \textbf{Output:} $V^c$.
\end{algorithmic}
\end{algorithm}

\begin{figure}[t]
\center
{\tiny
\begin{tabular}{ccc}
\multirow{1}{*}{\includegraphics[width=0.301\columnwidth]{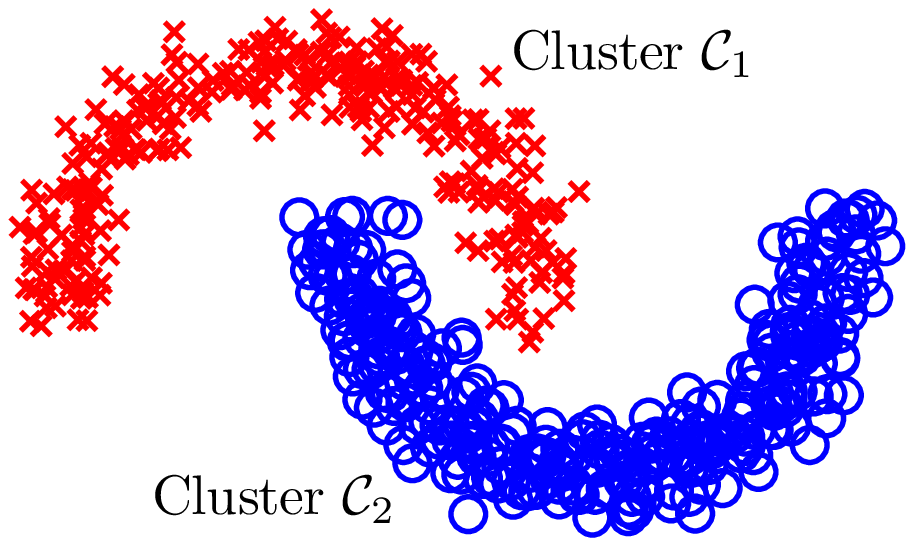}}
& 
\begin{tabular}[t]{ccc}
\rowcolor{orangecolor1} \rule{0pt}{.431cm} & Mean & PNZ \\
\rowcolor{orangecolor2} \rule{0pt}{.431cm} $\deg_i^-(\cluster)$ & 0.73 & 0.85 \\
\rowcolor{orangecolor3} \rule{0pt}{.431cm} $\deg_i^+(\cluster)$ & 0.94 & 0.54 \\
\rowcolor{orangecolor2} \rule{0pt}{.431cm} $\sqrt{\deg_i^-(\cluster)\deg_i^+(\cluster)}$ & $\mathbf{0.47}$ & $\mathbf{0.38}$
\end{tabular}
& 
\begin{tabular}[t]{ccc}
\rowcolor{orangecolor1} \rule{0pt}{.431cm} & Mean & PNZ \\
\rowcolor{orangecolor2} \rule{0pt}{.431cm} $\deg_i^-(\cluster)$ & 0.87 & 1.00 \\
\rowcolor{orangecolor3} \rule{0pt}{.431cm} $\deg_i^+(\cluster)$ & 0.87 & 1.00 \\
\rowcolor{orangecolor2} \rule{0pt}{.431cm} $\sqrt{\deg_i^-(\cluster)\deg_i^+(\cluster)}$ & $\mathbf{0.86}$ & $\mathbf{1.00}$ \\
\end{tabular} \\
 \rule{0pt}{.431cm} & $\mbox{Correlation}\left(\deg_i^-(\cluster), \deg_i^+(\cluster)\right) = 0.02$ & $\mbox{Correlation}\left(\deg_i^-(\cluster), \deg_i^+(\cluster)\right) = 0.80$ \\
 \rule{0pt}{.431cm} Synthetic data & (a) $i \in \cluster_1, \cluster = \cluster_2$ & (b) $i \in \cluster_1, \cluster = \cluster_1$
\end{tabular}
}
\caption{To verify the robustness of the product of indegree and outdegree as an affinity measure from a vertex $i$ to a cluster $\cluster$, we compare statistics in two cases: $i$ and $\cluster$ belong to different ground-truth clusters, e.g., $i \in \cluster_1$ and $\cluster = \cluster_2$ as in (a), and $i$ and $\cluster$ are in the same ground-truth cluster, e.g.,  $i \in \cluster_1$ and $\cluster = \cluster_1$ as in (b). We see that, in the first case, the product is a quantity more robust than the indegree or outdegree. For all $i\in\cluster_1$, such that $\deg_i^- > 0$ or $\deg_i^+ > 0$, the mean and proportion of nonzero values (PNZ) of $\sqrt{\deg_i^- \deg_i^+}$ are much smaller than those of $\deg_i^-$ and $\deg_i^+$, which implies a small affinity between $i$ and $\cluster$. Here the square root is for fair comparison of the quantities. In contrast, in the second case, the mean and PNZ of $\sqrt{\deg_i^- \deg_i^+}$ are close to those of $\deg_i^-$ and $\deg_i^+$, which means that the product keeps the large affinity well. The correlation of $\deg_i^-$ and $\deg_i^+$ , which is weak in (a) and strong in (b), further verifies the effectiveness of our affinity measure for reducing noisy edges across ground-truth clusters and keeping edges inside ground-truth clusters.} \label{fig:product}
\end{figure}

\subsection{Affinity Measure via Product of Indegree and Outdegree} \label{sec:affinity}
The affinity measure between two clusters is the key of an agglomerative clustering algorithm. Our affinity measure is based on \emph{indegree} and \emph{outdegree} in the graph representation.
For simplicity, we start from measuring the affinity between a vertex and a cluster.

\textbf{Indegree and outdegree}.
Considering a vertex and a cluster, the connectivity between them by inedges and outedges can be quantified using the concepts of indegree and outdegree.
\begin{defn}
Given a vertex $i$, the average indegree from and the average outdegree to a cluster $\mathcal{C}$ is defined as
$\deg_i^-(\mathcal{C}) = \frac{1}{|\mathcal{C}|} \sum_{j \in \mathcal{C}} w_{ji}$ and $\deg_i^+(\mathcal{C}) = \frac{1}{|\mathcal{C}|} \sum_{j \in \mathcal{C}} w_{ij}$, respectively, where $|\cluster|$ is the cardinality of set $\cluster$.
\end{defn}
As we stated in Sec. \ref{sec:intro}, the indegree measures the density near sample $i$, and the outdegree characterizes the $K$-NN similarity from vertex $i$ to cluster $\mathcal{C}$. We use the size of the cluster to normalize the degrees, otherwise, the algorithm may favor of merging large clusters instead of merging small clusters with dense connections. We find that in practice the normalized degrees work much better than the unnormalized degrees.

\textbf{Affinity between a vertex and a cluster}. A vertex should be merged to a cluster if it is strongly connected to the cluster by both inedges and outedges. Mathematically, the correlation of two types of degree is weak, if the vertex and the cluster belong to different ground-truth clusters, and strong, otherwise. To verify this intuition, we show such statistics on synthetic data in Fig. \ref{fig:product}. Therefore, we define the affinity as the product of the average indegree and average outdegree, i.e.,
\begin{equation}
\mathcal{A}_{i \rightarrow \mathcal{C}} = \deg_i^-(\mathcal{C}) \deg_i^+(\mathcal{C}).
\end{equation}
This affinity is robust to noisy edges between different ground-truth clusters because the product can be zero if the inedges and outedges do not coincide.

\textbf{Affinity between two clusters}. Following the above, we define the asymmetric affinity from cluster $\cluster_b$ to cluster $\cluster_a$ by summing up with respect to all the vertices in $\cluster_b$, i.e.,
\begin{equation} \label{eqn:diaffdef}
\mathcal{A}_{\mathcal{C}_b \rightarrow \mathcal{C}_a} = \sum_{i \in \mathcal{C}_b} \mathcal{A}_{i \rightarrow \mathcal{C}_a} = \sum_{i \in \mathcal{C}_b} \deg_i^-(\mathcal{C}_a) \deg_i^+(\mathcal{C}_a).
\end{equation}

Finally, we have the symmetric affinity used in our algorithm as
\begin{equation} \label{eqn:affdef}
\mathcal{A}_{\mathcal{C}_a, \mathcal{C}_b} = \mathcal{A}_{\mathcal{C}_b \rightarrow \mathcal{C}_a} + \mathcal{A}_{\mathcal{C}_a \rightarrow \mathcal{C}_b}
\end{equation}

\textbf{Efficient computation of affinity}. Our affinity measure can be computed efficiently using the following theorem.
\begin{thm} \label{thm:affinity}
The affinity between $\cluster_a$ and $\cluster_b$ defined in Eq. (\ref{eqn:affdef}) can be expressed in the matrix form
\begin{equation}  \label{eqn:aff}
\begin{array}{rcl}
\displaystyle \mathcal{A}_{\mathcal{C}_a, \mathcal{C}_b} & \displaystyle = & \displaystyle \frac{1}{|\mathcal{C}_a|^2} \mathbf{1}_{|\mathcal{C}_a|}^T \mathbf{W}_{\mathcal{C}_a, \mathcal{C}_b} \mathbf{W}_{\mathcal{C}_b, \mathcal{C}_a} \mathbf{1}_{|\mathcal{C}_a|}
 + \frac{1}{|\mathcal{C}_b|^2} \mathbf{1}_{|\mathcal{C}_b|}^T \mathbf{W}_{\mathcal{C}_b, \mathcal{C}_a} \mathbf{W}_{\mathcal{C}_a, \mathcal{C}_b} \mathbf{1}_{|\mathcal{C}_b|},
\end{array}
\end{equation}
where $\affm_{\mathcal{C}_a, \mathcal{C}_b}$ is the submatrix of $\affm$ whose row indices correspond to the vertices in $\cluster_a$ and column indices correspond to the vertices in $\cluster_b$, i.e., the weights of edges from $\cluster_a$ to $\cluster_b$, and $\mathbf{1}_{L}$ is an all-one vector of length $L$.
\end{thm}
\begin{rem}
The computation is reduced to vector additions and inner-products. So, our algorithm is easy to implement.
\end{rem}

\begin{proof}
It is easy to see that
\begin{equation}
\displaystyle \deg_i^-(\mathcal{C}_a) = \frac{1}{|\mathcal{C}_a|} \left[\mathbf{1}_{|\mathcal{C}_a|}^T \mathbf{W}_{\mathcal{C}_a, \mathcal{C}_b}\right]_i,
\end{equation}
\begin{equation}
\displaystyle \deg_i^+(\mathcal{C}_a) = \frac{1}{|\mathcal{C}_a|} \left[\mathbf{W}_{\mathcal{C}_b, \mathcal{C}_a} \mathbf{1}_{|\mathcal{C}_a|}\right]_i,
\end{equation}
where $\left[\mathbf{v}\right]_i$ is the $i$-th element of vector $\mathbf{v}$.
Then, by Eq. (\ref{eqn:diaffdef}), we can obtain the following lemma.
\begin{lem} \label{lem:directed_affinity}
\begin{equation} \label{eqn:directed_affinity}
\displaystyle \mathcal{A}_{\mathcal{C}_b \rightarrow \mathcal{C}_a} = \frac{1}{|\mathcal{C}_a|^2} \mathbf{1}_{|\mathcal{C}_a|}^T \mathbf{W}_{\mathcal{C}_a, \mathcal{C}_b} \mathbf{W}_{\mathcal{C}_b, \mathcal{C}_a} \mathbf{1}_{|\mathcal{C}_a|}.
\end{equation}
\end{lem}
Finally, Theorem \ref{thm:affinity} can be directly implied by Lemma \ref{lem:directed_affinity} using Eq. (\ref{eqn:affdef}).
\end{proof}

\textbf{Comparison to average linkage}. The GDL algorithm is different from average linkage in the following three aspects. First of all, the conventional average linkage is based on pairwise distances \cite{hastie2009elements}. Although we find that average linkage has much better performance on the $K$-NN graph than pairwise distances, we are unaware of any literature which studied the graph-based average linkage algorithm. Second, graph-based average linkage simply symmetrizes the directed graph by setting $w_{ij} = w_{ji} = (w_{ij} + w_{ji}) / 2$, while our algorithm uses the directed graph. Third, graph-based average linkage can be interpreted as defining the affinity measure $\mathcal{A}_{\mathcal{C}_b \rightarrow \mathcal{C}_a} = \frac{1}{|\mathcal{C}_b|} \sum_{i \in \mathcal{C}_b} [\deg_i^-(\mathcal{C}_a) + \deg_i^+(\mathcal{C}_a)]/2$ using our indegree-outdegree framework. The sum of the indegree and outdegree is not as robust as the product of them to noise. Experimental results in Fig. \ref{fig:toyexp} and Sec. \ref{sec:exp_img} demonstrate the superiority of GDL to graph-based average linkage.

\subsection{Implementations of GDL}
We present two implementations of the GDL algorithm: an exact algorithm via an efficient update formula and an approximate algorithm called Accelerated GDL (AGDL). Both implementations have the time complexity of $O(n^2)$ (see Theorem \ref{thm:complexity}).

\textbf{Update formula}. In each iteration, we select two clusters $\cluster_a$ and $\cluster_b$ with the largest affinity and merge them as $\cluster_{ab} = \cluster_a \cup \cluster_b$. Then, we need to update the asymmetric affinity $\mathcal{A}_{\mathcal{C}_{ab} \rightarrow \mathcal{C}_c}$ and $\mathcal{A}_{\mathcal{C}_{c} \rightarrow \mathcal{C}_{ab}}$, for any other cluster $\cluster_c$.

Using Lemma \ref{lem:directed_affinity}, we find that $\mathcal{A}_{\mathcal{C}_{ab} \rightarrow \mathcal{C}_c}$ can be computed as follows.
\begin{equation} \label{eqn:lw}
\mathcal{A}_{\mathcal{C}_{ab} \rightarrow \mathcal{C}_c} = \mathcal{A}_{\mathcal{C}_{a} \rightarrow \mathcal{C}_c} + \mathcal{A}_{\mathcal{C}_{b} \rightarrow \mathcal{C}_c}.
\end{equation}
By storing all the asymmetric affinities, the update is simple.

As the same update formula cannot be applied to $\mathcal{A}_{\mathcal{C}_{c} \rightarrow \mathcal{C}_{ab}}$, we have to compute it directly using Eq. (\ref{eqn:directed_affinity}). However, the total complexity is $O(n)$ in each iteration, due to the row sparsity of $\mathbf{W}$ (see Sec. \ref{sec:compproof} in the supplemental materials for details).

The GDL algorithm with the update formula (GDL-U) is presented as Algorithm \ref{alg:gdllw} in the supplemental materials.

\textbf{Accelerated GDL}.
Although the GDL-U algorithm is simple and fast, we further propose AGDL. The major computational cost is on computing the affinities. To reduce the number of affinities computed in each iteration, AGDL maintains a neighbor set of size $K^c$ for each cluster in $V^c$, to approximate its $K^c$-nearest cluster set. Then, finding the maximum affinity among all pairs of clusters can then be approximated by searching it in all the neighbor sets. Updating the neighbor sets involves computation of the affinity between the new cluster and a small set of clusters, instead of all the other clusters.

Denote the neighbor set of a cluster $\cluster$ as $\mathcal{N}_\cluster$. Initially $\mathcal{N}_\cluster$ consists of $\cluster$'s $K^c$-nearest clusters. Once two clusters $\cluster_a$ and $\cluster_b$ are merged, we need to update the neighbor sets which include $\cluster_a$ or $\cluster_b$, and create the neighbor set of $\cluster_a \cup \cluster_b$. We utilize two assumptions that (1)  if $\cluster_a$ or $\cluster_b$ is among the $K^c$-nearest clusters of $\cluster_c$, $\cluster_a \cup \cluster_b$ is probably among the $K^c$-nearest clusters of $\cluster_c$; (2)  if $\cluster_c$ is among the $K^c$-nearest clusters of $\cluster_a$ or $\cluster_b$, $\cluster_c$ is probably among the $K^c$-nearest clusters of $\cluster_a \cup \cluster_b$. So, the new cluster $\cluster_a \cup \cluster_b$ is added to the neighbor sets which include $\cluster_a$ or $\cluster_b$ previously. To create the neighbor set for $\cluster_a \cup \cluster_b$, we select the $K^c$-nearest clusters from $\mathcal{N}_{\cluster_a} \cup \mathcal{N}_{\cluster_b}$.

The AGDL algorithm is summarized in Algorithm \ref{alg:agdl} in the supplemental materials.

\subsection{Time Complexity Analysis}
We have the following theorem about the time complexity of the GDL, GDL-U and AGDL algorithms (please refer to Sec. \ref{sec:compproof} in the supplemental materials for the proof).
\begin{thm} \label{thm:complexity}~
\begin{enumerate}[(a)]
\item The time complexity of the GDL algorithm (i.e., Algorithm \ref{alg:gdl}) is $O(n^3)$.
\item The time complexity of the GDL-U algorithm (i.e., Algorithm \ref{alg:gdllw}) is $O(n^2)$.
\item The time complexity of the AGDL algorithm (i.e., Algorithm \ref{alg:agdl}) is $O(n^2)$.
\end{enumerate}
\end{thm}

\section{Experiments} \label{sec:exp}
In this section, we demonstrate the effectiveness of GDL and AGDL on image clustering and object matching.
All the experiments are run in MATLAB on a PC with 3.20GHz CPU and 8G memory.

\subsection{Image Clustering} \label{sec:exp_img}
We carry out experiments on six publicly available image benchmarks, including object image databases (COIL-20 and COIL-100), hand-written digit databases (MNIST and USPS), and facial image databases (Extended Yale-B, FRGC ver2.0).\footnote{COIL-20 and COIL-100 are from \url{http://www.cs.columbia.edu/CAVE/software/}. MNIST and USPS are from \url{http://www.cs.nyu.edu/~roweis/data.html}. Extended Yale-B is from \url{http://vision.ucsd.edu/~leekc/ExtYaleDatabase/ExtYaleB.html}. FRGC ver2.0 is from \url{http://face.nist.gov/frgc/}.} For MNIST, we use all the images in the testing set. For FRGC ver2.0, we use all the facial images in the training set of experiment 4. The statistics of all the datasets are presented in Table \ref{tab:imgtime}. We adopt widely used features for different kinds of images: the intensities of pixels as features and Euclidean distance for object and digit images, and local binary patterns (LBP) as features and $\chi^2$ distance for facial images.


We compare the GDL-U and AGDL with eight representative algorithms, i.e., $k$-medoids ($k$-med) \cite{hastie2009elements}, average linkage (Link) \cite{hastie2009elements}, graph-based average linkage (G-Link), normalized cuts (NCuts) \cite{shi2000ncut}\footnote{The code is downloaded from \url{http://www.cis.upenn.edu/~jshi/software/}, which implements the multiclass normalized cuts algorithm \cite{yu2003multiclass}.}, NJW spectral clustering (NJW-SC) \cite{ng2001spectral}, directed graph spectral clustering (DGSC) \cite{zhou2005directed}, self-tuning spectral clustering (STSC) \cite{zelnik2004self} and Zell \cite{zhao2008zeta}. Here we use $k$-medoids instead of $k$-means because it can handle the case where distances between points are not measured by Euclidean distances. To fairly compare the graph-based algorithms, we fix $K = 20$ and select $a$ with the best performance from the set $\{10^i, i \in [-2:0.5:2]\}$ on all the datasets. For our algorithms, the parameters are fixed as $K^0 = 1$, $K^c = 10$. The numbers of ground-truth clusters are used as the input of all algorithms (e.g., $n_T$ in our algorithm).

We adopt the widely used Normalized Mutual Information (NMI) \cite{wu2007local} to quantitatively evaluate the performance of clustering algorithms. The NMI quantifies the normalized statistical information shared between two distributions. A larger NMI value indicates a better clustering result.

The results measured in NMI are given in Table \ref{tab:imgnmi}. $k$-medoids and average linkage perform similar, as they heavily rely on the computation of pairwise distances and thus are sensitive to noise, and cannot well capture the complex cluster structures in the real data sets. NCuts, NJW-SC, and Zell have good performance on most data sets, as they capture the underlying manifold structures of the data. STSC works fine on some synthetic multiscale datasets in \cite{zelnik2004self} but its results are worse than ours on several real datasets in comparison. Note that STSC adaptively estimated the parameter $\sigma^2$ at every point to reflect the variation of local density while ours explores indgree/outdegree and fixes $\sigma^2$ as constant. The effective and robust affinity measure for agglomerative clustering makes our GDL-U and AGDL algorithm performs the best among all the algorithms. The AGDL's results are nearly the same as GDL-U.

\begin{figure}[t]
\center
\begin{tabular}{c@{\extracolsep{2em}}c}
\includegraphics[width=0.366\columnwidth]{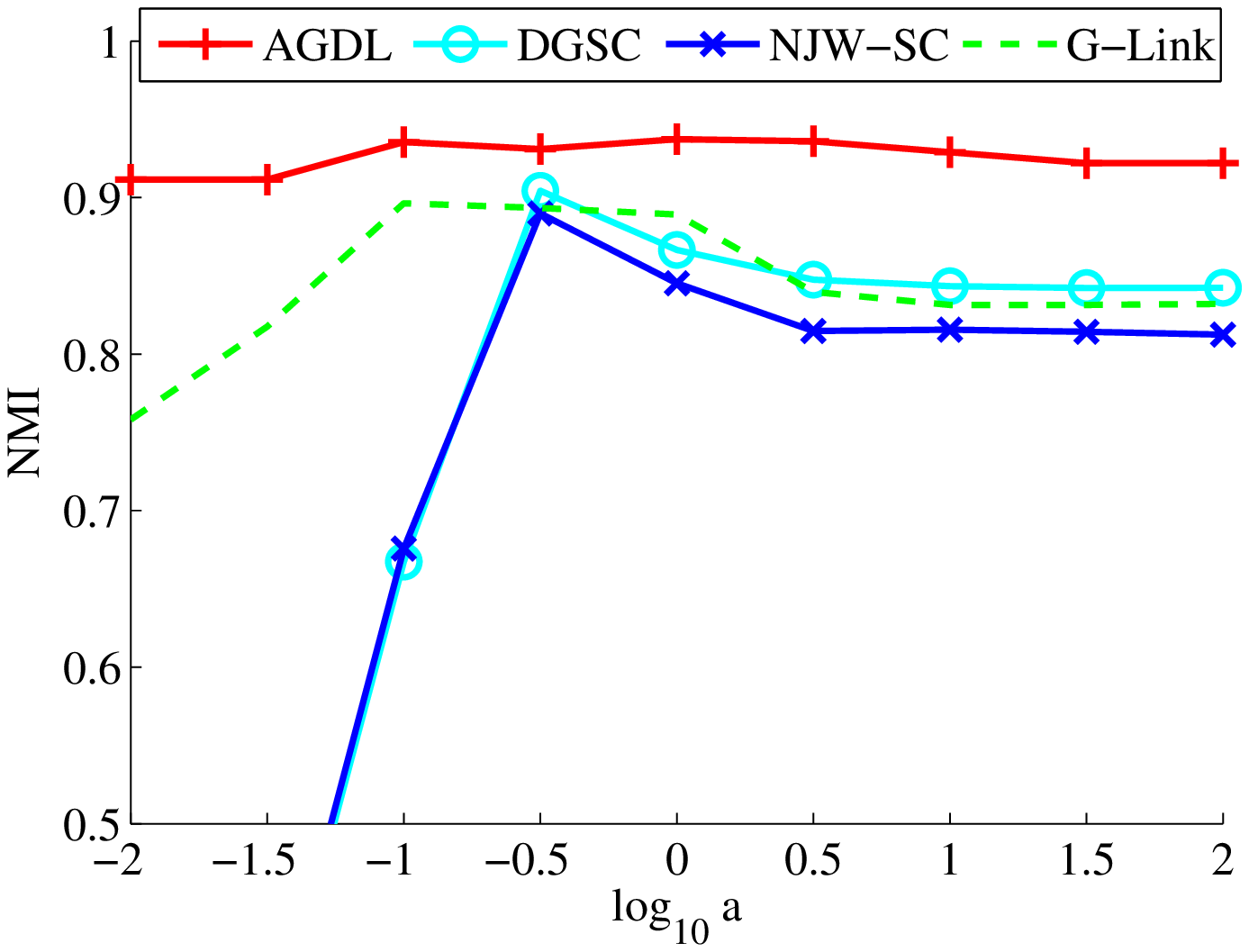}
& \includegraphics[width=0.366\columnwidth]{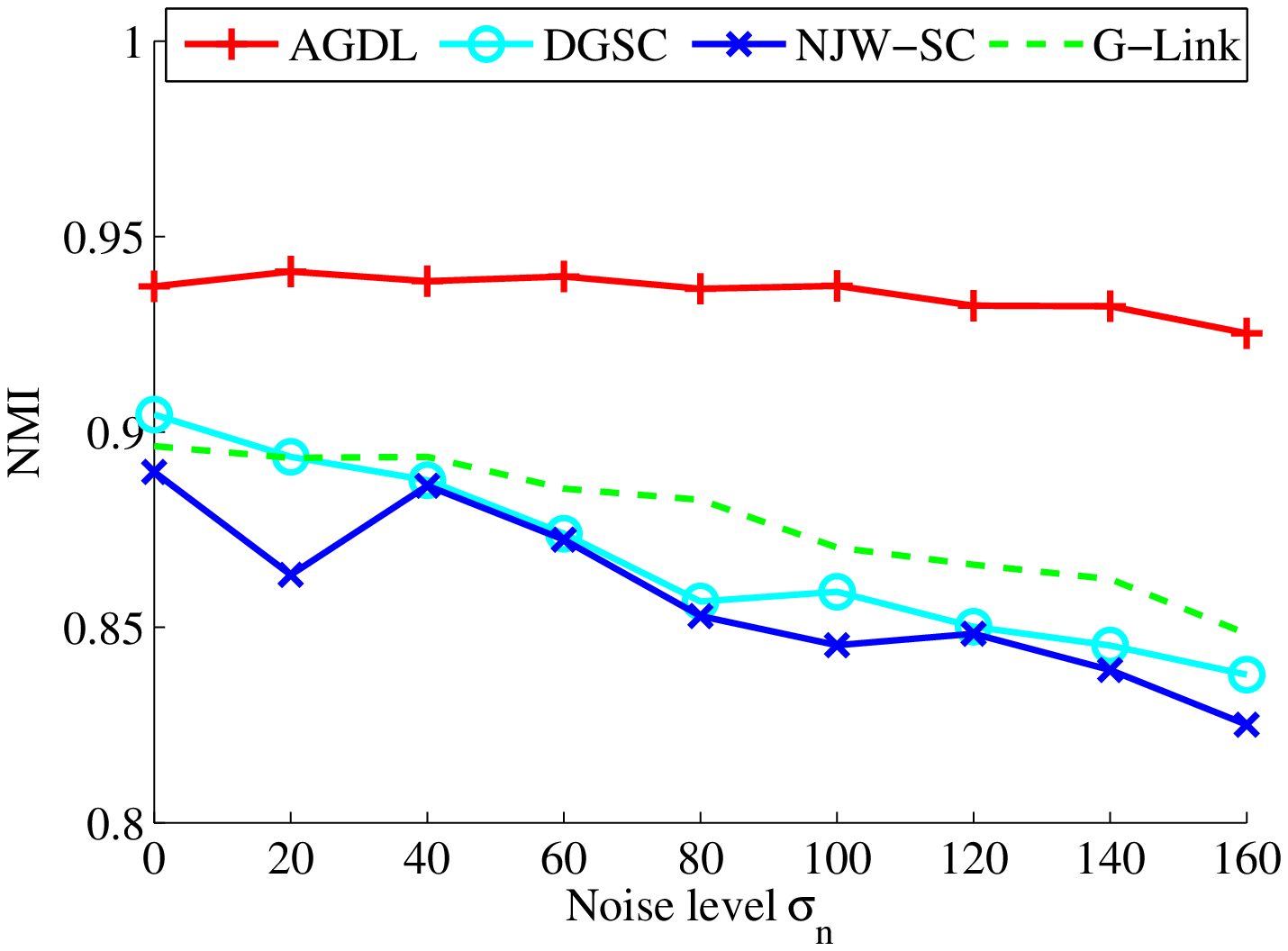} \\
 (a) & (b)
\end{tabular}
\caption{Variations of performance of different clustering algorithms on the COIL-20 dataset, (a) when the parameter $a$ for controlling $\sigma$ in Eq. (\ref{eqn:graphw}) changes; (b) when we add Gaussian noise $\mathcal{N}(0, \sigma_n^2)$ to the images. The NMI differences between $\sigma_n = 0$ and $\sigma_n = 160$ are $0.048$, $0.065$, $0.067$, $0.012$, for G-Link, NJW-SC, DGSC, and AGDL, respectively.
} \label{fig:clusterparam}
\end{figure}

Compared to other graph-based algorithms, GDL-U and AGDL are more robust to the parameter $\sigma$ for building the graph, as well as the noise in the data (see Fig. \ref{fig:clusterparam}). The noise added to images can degrade the performance of other algorithms greatly, but our performance is barely affected.


\begin{table}[t]
\center
\caption{Quantitative clustering results in NMI on real imagery data. A larger NMI value indicates a better clustering result. The results shown in a boldface are significantly better than the others, with a significance level of 0.01.} \label{tab:imgnmi}
{
\begin{tabular}{|l||c|c|c|c|c|c|c|c||c|c|}
\hline {Dataset} & k-med & Link  & G-Link & NCuts & NJW-SC & DGSC  & STSC  &  Zell & GDL-U & AGDL \\
\hline
COIL-20          & 0.710 & 0.647 &  0.896 & 0.884 &  0.889 & 0.904 & 0.895 & 0.911 & \textbf{0.937} & \textbf{0.937} \\
COIL-100         & 0.706 & 0.606 &  0.855 & 0.823 &  0.854 & 0.858 & 0.858 & 0.913 & \textbf{0.929} & \textbf{0.933} \\
{USPS}           & 0.336 & 0.095 &  0.732 & 0.675 &  0.690 & 0.747 & 0.726 & 0.799 & \textbf{0.824} & \textbf{0.824} \\
{MNIST}          & 0.390 & 0.304 &  0.808 & 0.753 &  0.755 & 0.795 & 0.756 & 0.768 & \textbf{0.844} & \textbf{0.844} \\
Yale-B           & 0.329 & 0.255 &  0.766 & 0.809 &  0.851 & 0.869 & 0.860 & 0.781 & \textbf{0.910} & \textbf{0.910} \\
FRGC             & 0.541 & 0.570 &  0.669 & 0.720 &  0.723 & 0.732 & 0.729 & 0.653 & \textbf{0.747} & \textbf{0.746} \\
\hline
\end{tabular}
}
\end{table}

\begin{table}[t]
\center
\caption{The time cost (in \emph{seconds}) of the algorithms. The minimum time cost on each dataset is in bold. The statistics of each dataset are shown for reference.} \label{tab:imgtime}
{
\begin{tabular}{|l||c|c||c|c|c|c|c|c|c|}
\hline
{Dataset} & Sample Num & Cluster Num & {NCuts} & {NJW-SC} & DGSC & {Zell} & {GDL-U} & {AGDL} \\ 
\hline
COIL-20     &   1440   &    20      & 3.880  & 6.399   &  8.280  & 15.22 &  \textbf{0.265}  & 0.277     \\ 
COIL-100    &   7200   &    100     & 133.8  & 239.7   &  326.4  & 432.9 &  12.81  & \textbf{5.530}     \\ 
{USPS}      &   11000  &    10      & 263.0  & 461.6   &  538.9  & 9703  &  53.64  & \textbf{29.01}     \\ 
{MNIST}     &   10000  &    10      & 247.2  & 384.4   &  460.4  & 64003 &  35.60  & \textbf{17.18}     \\   
Yale-B      &   2414   &    38      & 9.412  & 13.99   &  16.00  & 178.2 &  0.731  & \textbf{0.564}     \\ 
FRGC        &   12776  &    222     & 577.4  & 914.3   &  1012.2 & 65021 &  49.15  & \textbf{18.62} \\ 
\hline
\end{tabular}
}
\end{table}

For the graph-based algorithms, we show their time cost in Table \ref{tab:imgtime}. AGDL costs the least amount of time among all the algorithms. GDL is faster than NCuts, NJW-SC, and DGSC, and is much faster than Zell. G-Link, which has worse performance than AGDL, is comparable to AGDL on time cost.


\begin{figure}[t]
\center
{\footnotesize
\begin{tabular}{lll}
\includegraphics[width=0.316\columnwidth]{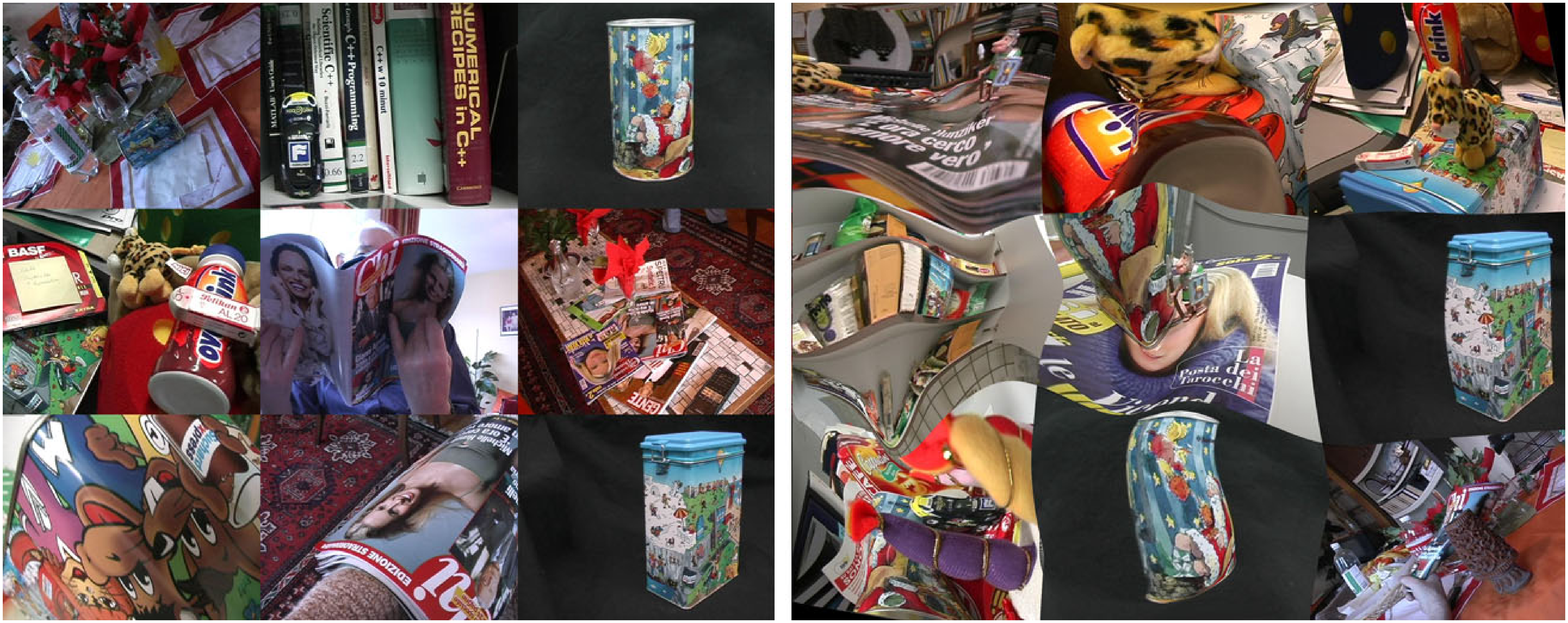}
& \includegraphics[width=0.310\columnwidth]{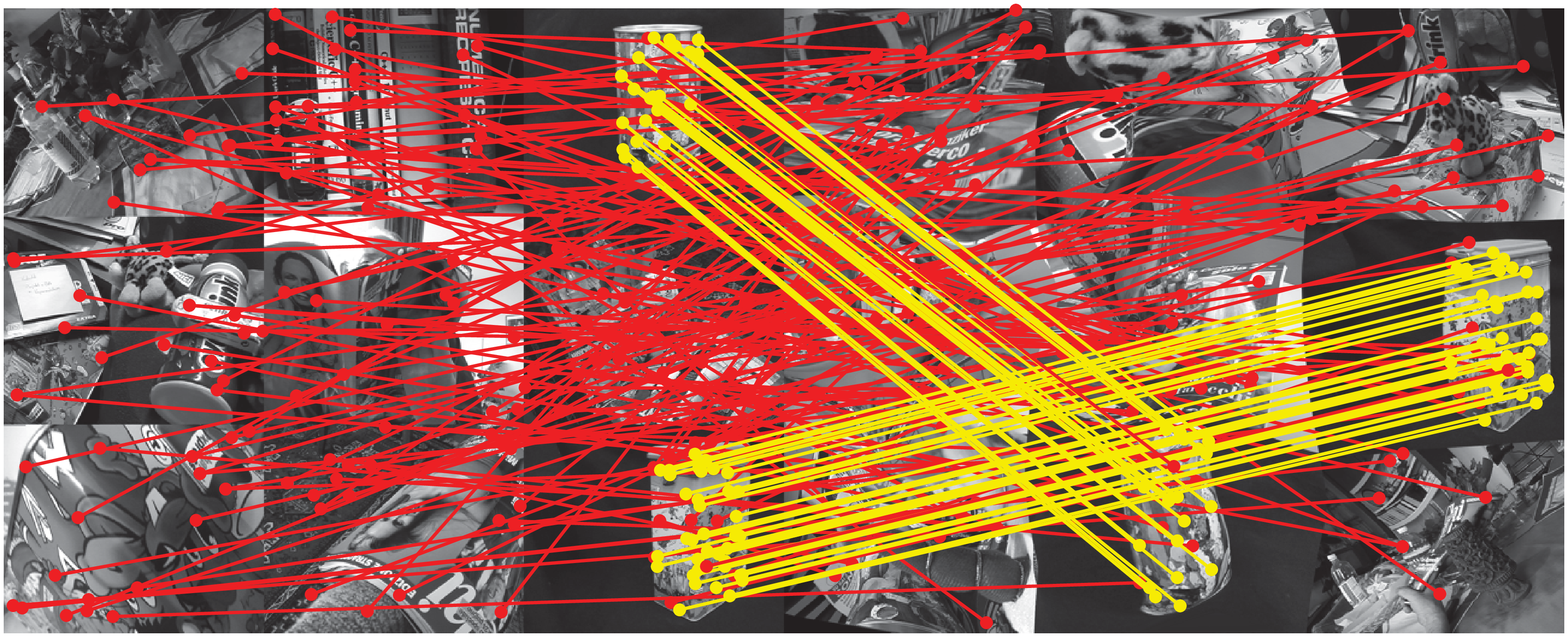}
& \includegraphics[width=0.310\columnwidth]{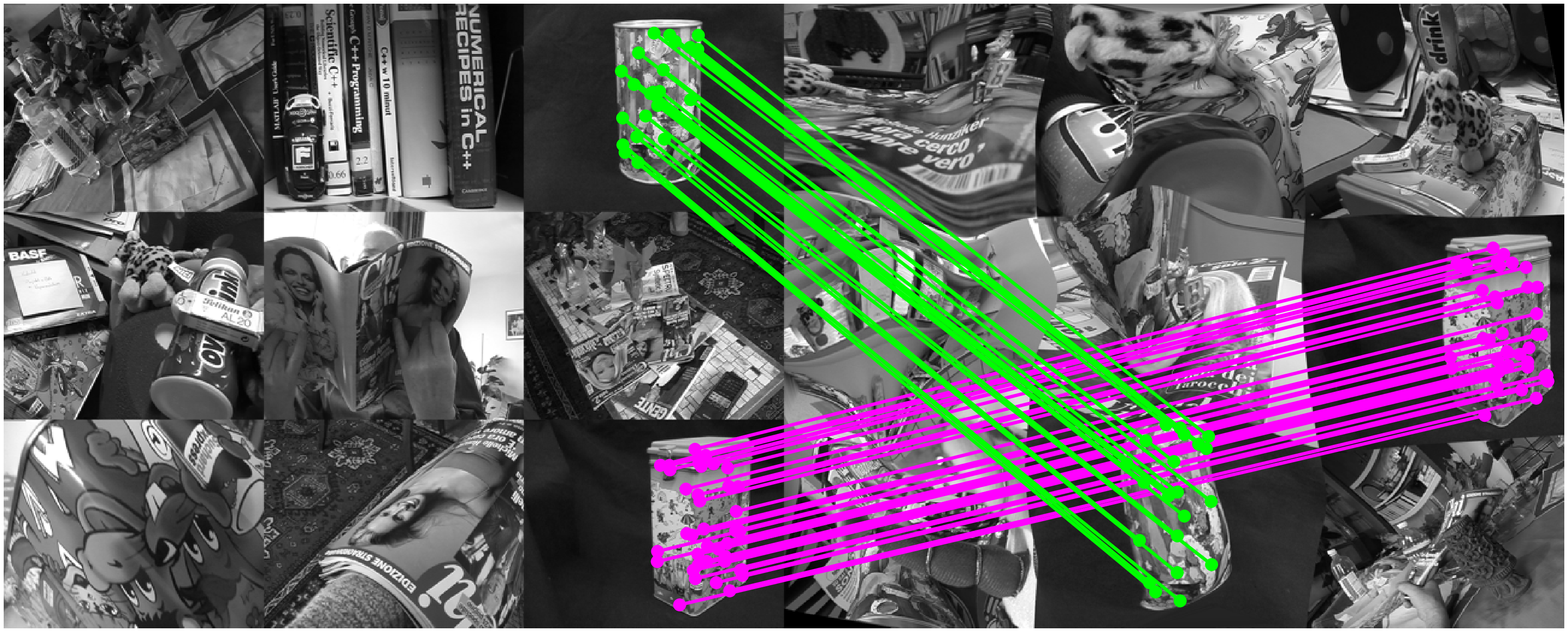} \\
 (a) A pair of composite images,
& (b) Initial correspondences
& (c) Detected correspondences \\
  one of which is warped with
& ($533$ inliers in yellow color,
& by AGDL ($532$ true, $552$  \\
 $\sigma_n = 50$; & $1200$ outliers in red color,  &  detected, F-score $0.981$). \\
 & according to the ground truth); &
\end{tabular}
}
\caption{Example of object matching through feature correspondence clustering.} \label{fig:match_example}
\end{figure}

\begin{figure}[t]
\center
\begin{tabular}{ccc}
\includegraphics[width=0.316\columnwidth]{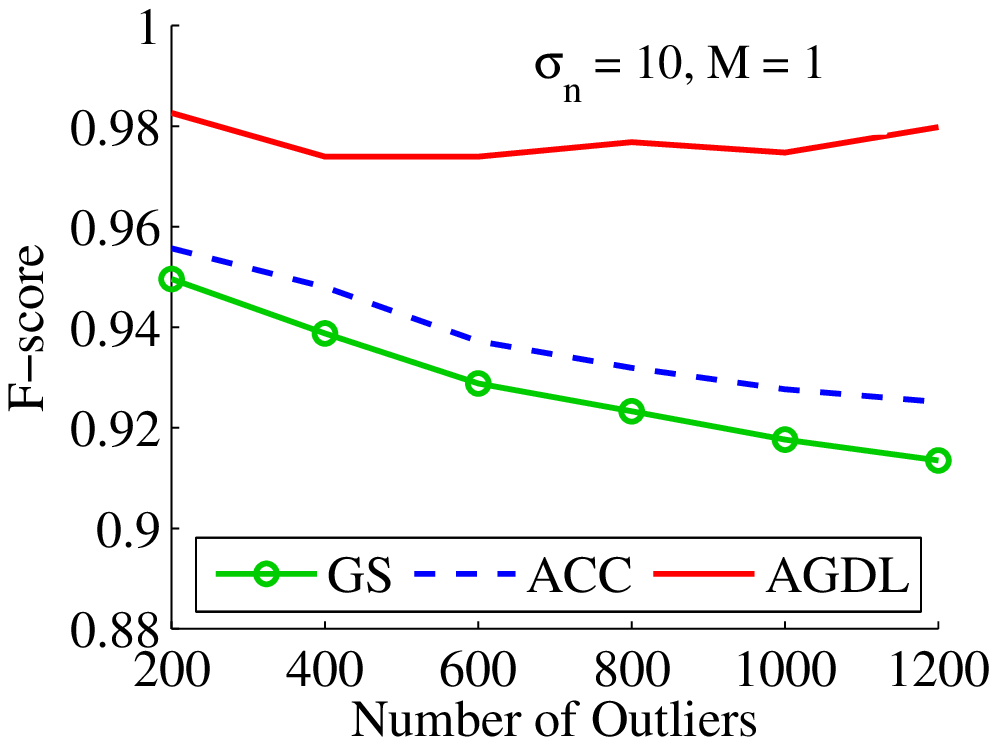}
& \includegraphics[width=0.316\columnwidth]{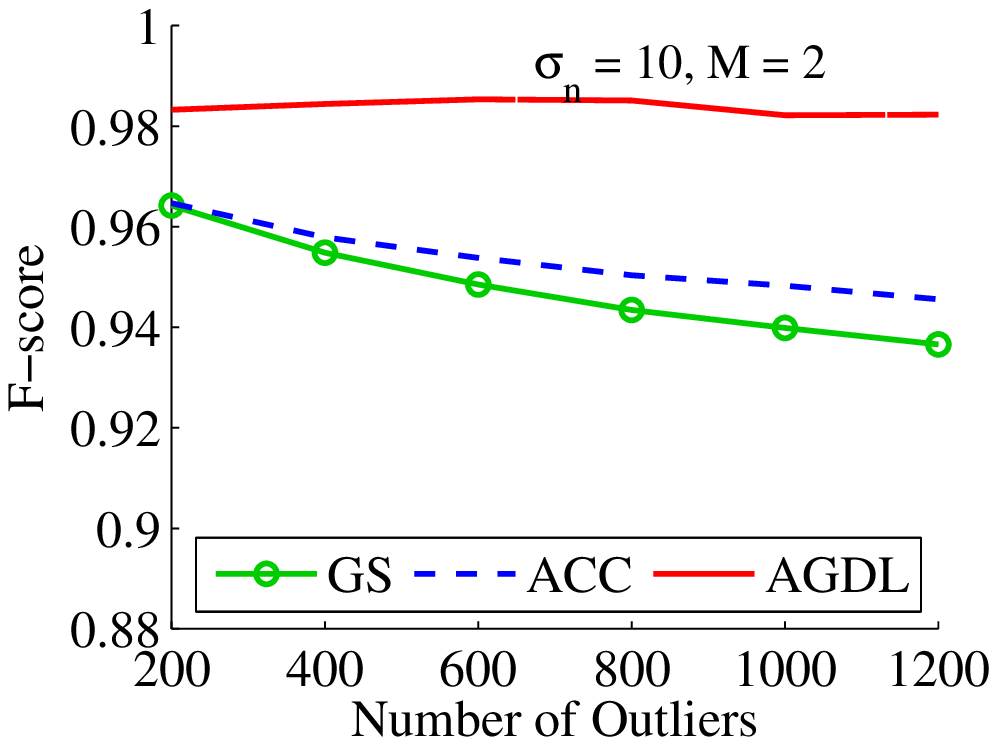}
& \includegraphics[width=0.316\columnwidth]{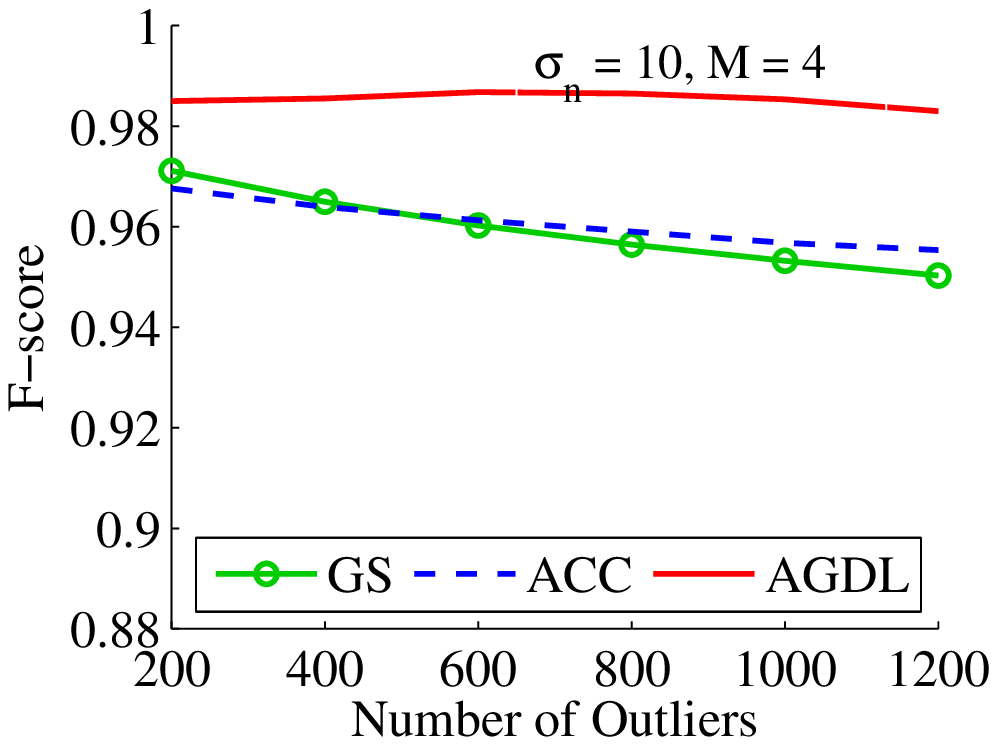} \\
\includegraphics[width=0.316\columnwidth]{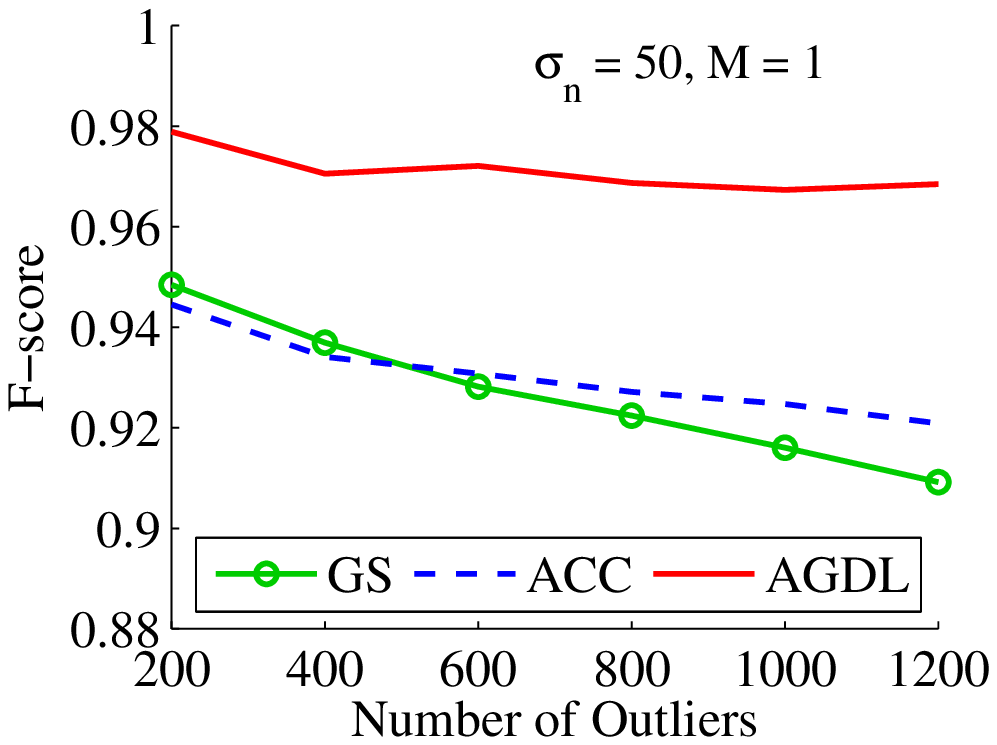}
& \includegraphics[width=0.316\columnwidth]{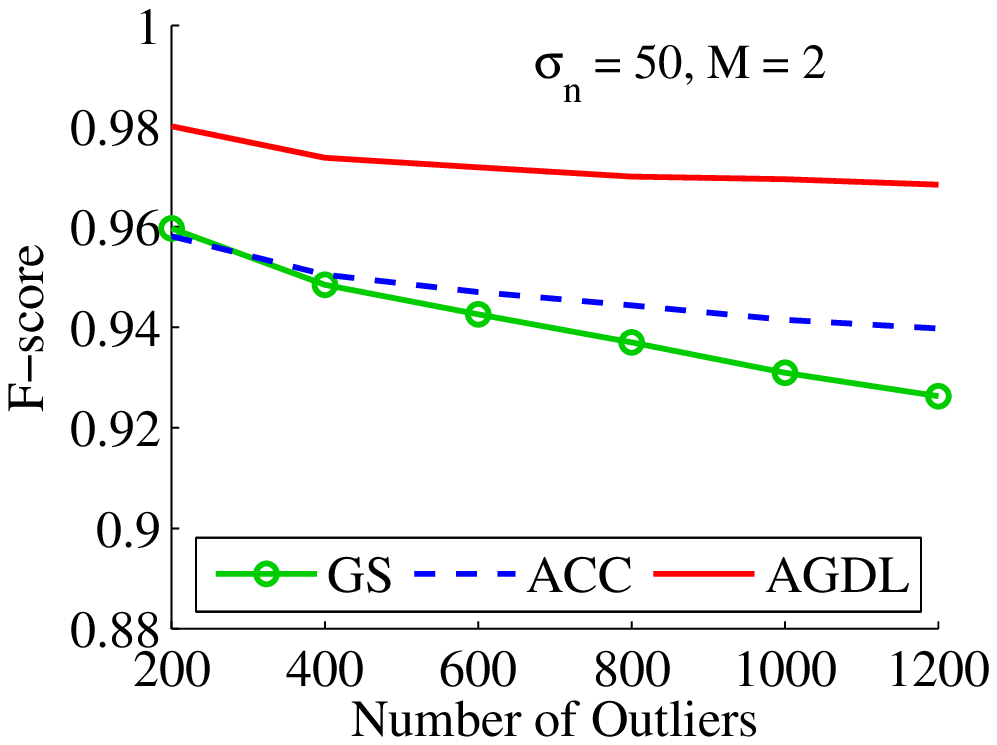}
& \includegraphics[width=0.316\columnwidth]{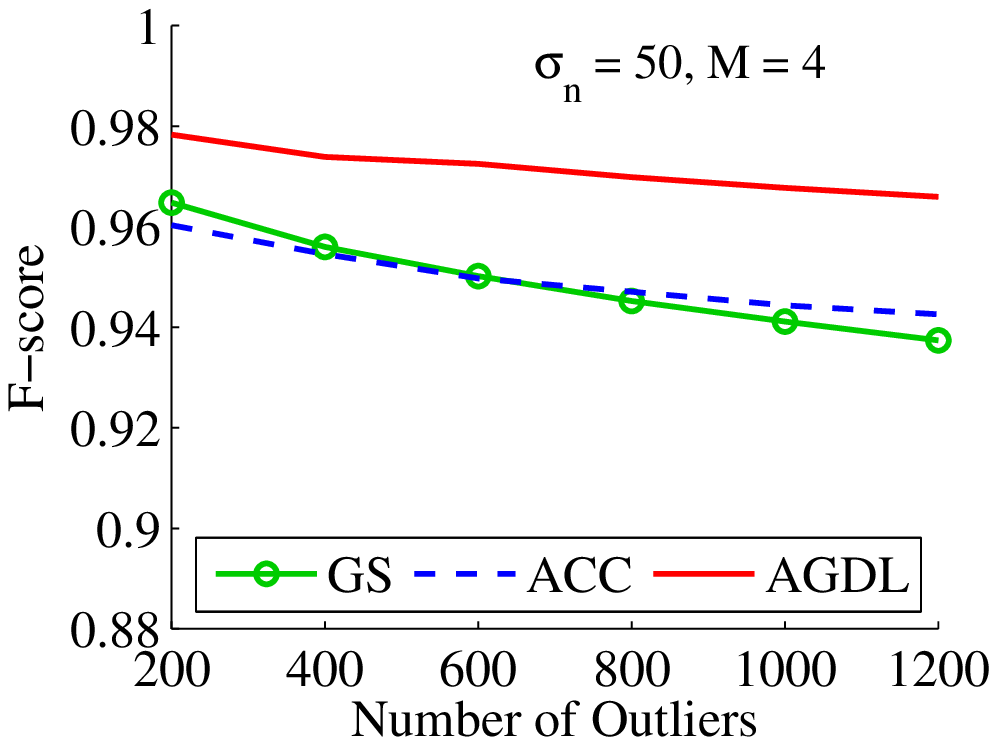} \\
\includegraphics[width=0.316\columnwidth]{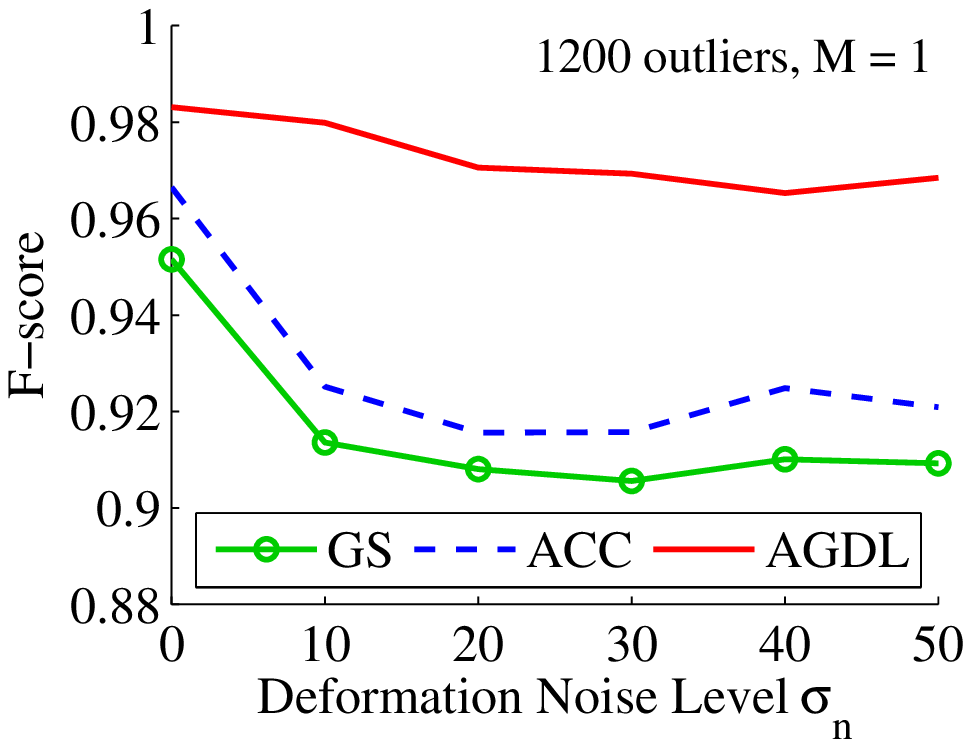}
& \includegraphics[width=0.316\columnwidth]{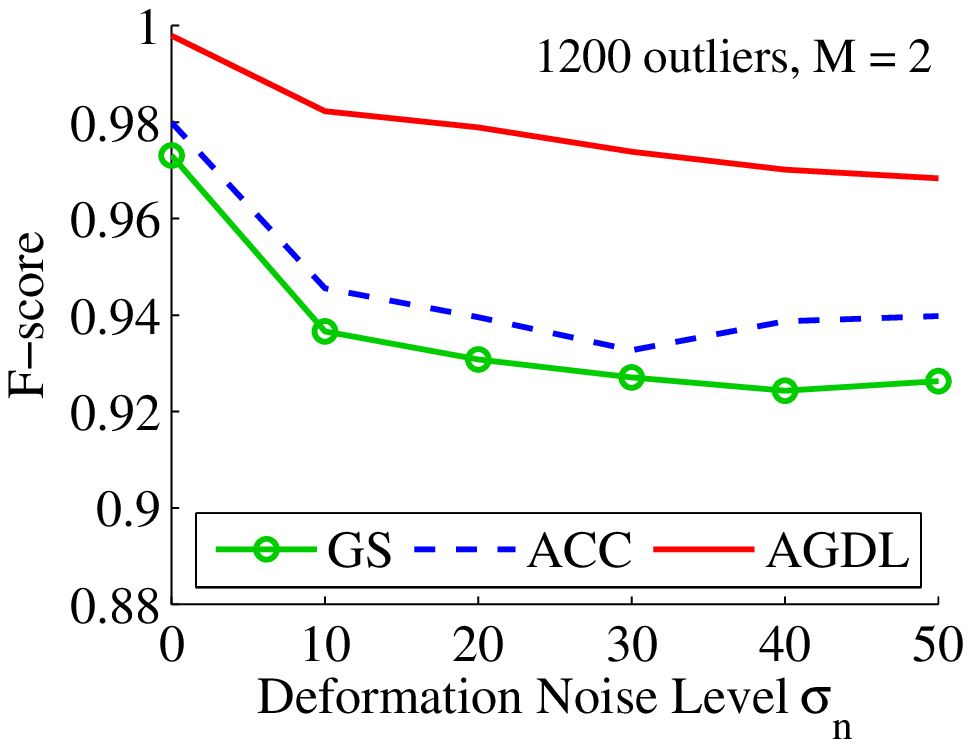}
& \includegraphics[width=0.316\columnwidth]{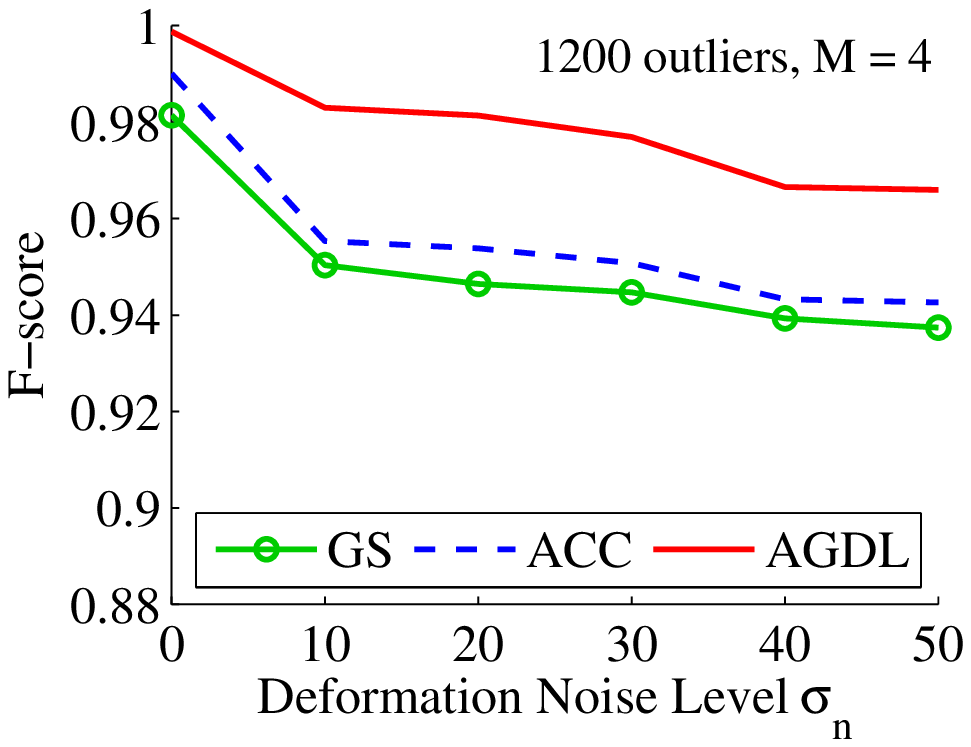}
\end{tabular}
\caption{Performance comparison of different algorithms. In each sub-figure, one of the three factors, i.e., the number of outliers, the level of deformation $\sigma_n$, and the number of common sub-images $M$, is varied, while the other two are fixed as the values appearing at the top. All the results are averaged over $30$ random trials.} \label{fig:img_match}
\end{figure}

\subsection{Feature Correspondence Clustering for Object Matching} \label{sec:img_match}
We show the effectiveness of our clustering algorithm in the presence of outliers via feature correspondence clustering.
Feature correspondence clustering is commonly used for robust object matching \cite{leordeanu2005spectral,cho2009feature,liu2010common}, which can deal with geometric distortions of objects across images and is a fundamental problem in computer vision. We demonstrate that our algorithm can be effectively integrated with the framework of feature correspondence clustering. Therefore, it has a range of potential applications, such as object recognition, image retrieval, and 3D reconstruction.

We compare with two recent state-of-the-art methods, i.e., agglomerative correspondence clustering (ACC) \cite{cho2009feature} and graph shift (GS) \cite{liu2010common}.\footnote{The code of ACC and GS are downloaded from \url{http://cv.snu.ac.kr/research/~acc/}, and \url{http://sites.google.com/site/lhrbss/}, respectively. We do not present the results of spectral matching (SM) \cite{leordeanu2005spectral}, because both ACC and GS outperformed SM greatly \cite{cho2009feature,liu2010common}, especially when there existed at least two clusters of correspondences according to the ground-truth.}


\textbf{Overview of experiments}. We follow the experiments in the ACC paper \cite{cho2009feature}. We use composite images and their warped versions (Fig. \ref{fig:match_example}(a)) to simulate cluttered scenes where deformable objects appear. Then we can use the ground-truth for performance evaluation. Namely, we compute the precision and recall rates of detected correspondences (Fig. \ref{fig:match_example}(c)), given a set of correspondences with ground-truth (Fig. \ref{fig:match_example}(b)). A good clustering algorithm can group inliers and separate outliers. It is a more direct way of evaluating the performance of clustering algorithms than other experiments, such as object recognition.

\textbf{Experimental settings}. We generate a pair of $3\times3$ tiled images that contain $M$ common sub-image(s). The common sub-images are randomly selected from the model images of the ETHZ toys dataset\footnote{\url{http://www.vision.ee.ethz.ch/~calvin/datasets.html}.}, and the non-common sub-images are from test images of the same dataset. The positions of all sub-images are randomly determined. When $M > 1$, the common sub-images are chosen as different objects. To simulate deformation, one of the paired images is warped using the thin-plate spline (TPS) model. An example of paired test images are shown in Fig. \ref{fig:match_example}(a). $9\times9$ crossing points from a $10\times10$ meshgrid on the image are chosen as the control points of the TPS model. Then, all the control points are perturbed by Gaussian noise of $N(0, \sigma_n^2)$ independently, and the TPS warping is applied based on the perturbations of control points. To obtain the candidate correspondences between two tiled images, features are extracted by the MSER detector, and the best $3,000$ correspondences are collected according to similarity of the SIFT descriptors. Using the warping model, each correspondence has a ground-truth label: true if its error is smaller than three pixels, and false otherwise. Fig. \ref{fig:match_example}(b) shows the correspondences as lines, among which the yellow ones represent true correspondences. Then, the performance of different algorithms are quantitatively evaluated. We use the F-score, a traditional statistical measure of accuracy, which is defined as $[\mbox{precision}\cdot\mbox{recall}/(\mbox{precision}+\mbox{recall})]$.

\textbf{Parameters of ACC and GS}. As we strictly follow the test protocol in the ACC paper \cite{cho2009feature}, we use the default parameters in their codes. For GS, we compute the affinity matrix $\mathbf{W}_{ij} = \max(\beta - d_{ij} / \sigma_s^2, 0)$ as the paper \cite{liu2010common}, where $d_{ij}$ is the distance between correspondence $i$ and correspondence $j$ as defined in the ACC paper \cite{cho2009feature}. $\beta$, $\sigma_s$ and other parameters in GS are tuned to be the best.

\textbf{Parameters of AGDL}. For our AGDL algorithm (i.e., Algorithm \ref{alg:agdl}), the parameters are fixed as $n_T = 50$, $a = 10$, $K = 35$, $K^0 = 2$, and $K^c = 10$. We found that the GDL works well in a large range of $n_T$, as the number of ground-truth clusters (i.e., $M$) is very small and we can eliminate the outlier clusters by postprocessing.\footnote{Please see Sec. \ref{sec:olagdl} in the supplemental materials for details of outlier elimination in our algorithm. Different from ACC, which utilizes \emph{additional} information, i.e., geometrical locations of feature points, we only use the $K$-NN graph in outlier elimination.}



\textbf{Results}. As shown in Fig. \ref{fig:img_match}, we vary the number of outliers, the level of deformation, and the number of common sub-images, and
compare the F-scores of detected correspondences by different algorithms. Both ACC and GS perform excellently on this task. It is challenging to beat them, which are very recent methods designed specifically for object matching. However, our simple clustering algorithm outperforms them. We find our AGDL algorithm performs consistently better than both ACC and GS under different settings. AGDL has a higher F-score than both in $95.6\%$ of the random trials under all the setting combinations. We attribute the success of AGDL to the effective cluster affinity measure which is robust to noise and outliers.



\section{Conclusion} \label{sec:con}
We present a fast and effective method for agglomerative clustering on a directed graph. Our algorithm is based on indegree and outdegree, fundamental concepts in graph theory. The indegree and outdegree have been widely studied in complex networks, but have not received much attention in clustering. We analyze their roles in modeling the structures of data, and show their power via the proposed graph degree linkage algorithm. We demonstrated the superiority of this simple algorithm on image clustering and object matching. We believe our work provides not only a simple and powerful clustering algorithm to many applications in computer vision, but also an insightful analysis of the graph representation of data via indegree and outdegree.


\section*{Acknowledgment}
{
This work is partially supported by the General Research Fund sponsored by the Research Grants Council of Hong Kong (Project No. CUHK416510, CUHK417110 and CUHK417011) and National Natural Science Foundation of China (Project No.61005057). It is also supported through Introduced Innovative R\&D Team of Guangdong Province 201001D0104648280 and Shenzhen Key Lab of Computer Vision and Pattern Recognition.
The authors would like to thank Tianfan Xue for proof reading and Wei Li for help on the ACC code.}

{\footnotesize
\bibliographystyle{splncs}
\bibliography{gdl}
}

\newpage
\section{Implementations of GDL}
\begin{algorithm}
\caption{Graph Degree Linkage with the update formula (GDL-U)} \label{alg:gdllw}
\begin{algorithmic}
\STATE \textbf{Input:} a set of $n$ samples $\mathcal{X} = \{\mathbf{x}_1, \mathbf{x}_2, \cdots, \mathbf{x}_n\}$, and the target number of clusters $n_T$.
\STATE Build the $K^0$-NN graph, and detect its weakly connected components as initial clusters. Denote the set of initial clusters as $V^c = \{\cluster_1, \cdots, \cluster_{n_c}\}$, where $n_c$ is the number of clusters.
\STATE Build the $K$-NN graph, and get the weighted adjacency matrix $\mathbf{W}$.
\STATE Initialize the asymmetric affinity table $\mathcal{A}_{\cluster_{a} \rightarrow \cluster_b}$ for $\cluster_{a}, \cluster_{b} \in V^c$.
\WHILE{$n_c > n_T$}
\STATE Search two clusters $\cluster_a$ and $\cluster_b$, such that $\{\cluster_a,\cluster_b\} = \operatornamewithlimits{argmax}_{\cluster_a, \cluster_b \in V^c} \mathcal{A}_{\cluster_a, \cluster_b}$;
\STATE $V^c \leftarrow \{V^c \setminus \{\cluster_a,\cluster_b\}\} \cup \{\cluster_{ab}\}$, where $\cluster_{ab} = \cluster_a \cup \cluster_b$, and $n_c = n_c - 1$;
\STATE For all $\cluster_c$, compute $\mathcal{A}_{\mathcal{C}_{ab} \rightarrow \mathcal{C}_c}$ using the update formula, i.e., Eq. (\ref{eqn:lw}), and $\mathcal{A}_{\mathcal{C}_{c} \rightarrow \mathcal{C}_{ab}}$ using Eq. (\ref{eqn:directed_affinity}).
\ENDWHILE
\STATE \textbf{Output:} $V^c$.
\end{algorithmic}
\end{algorithm}

\begin{algorithm}
\caption{Accelerated Graph Degree Linkage (AGDL)} \label{alg:agdl}
\begin{algorithmic}
\STATE \textbf{Input:} a set of $n$ sample vectors $\mathcal{X} = \{\mathbf{x}_1, \mathbf{x}_2, \cdots, \mathbf{x}_n\}$, and the target number of clusters $n_T$.
\STATE Build the $K^0$-NN graph, and detect its weakly connected components as initial clusters. Denote the set of initial clusters as $V^c = \{\cluster_1, \cdots, \cluster_{n_c}\}$, where $n_c$ is the number of clusters.
\STATE Build the $K$-NN graph, and get the weighted adjacency matrix $\mathbf{W}$.
\STATE Create a neighbor set for each cluster in $V^c$, and initialize it as the $K^c$-nearest cluster set.
\WHILE{$n_c > n_T$}
\STATE Search two clusters $\cluster_a$ and $\cluster_b$ from the affinity of pairs of clusters associated with the neighbor sets, such that $\{\cluster_a,\cluster_b\} = \operatornamewithlimits{argmax}_{\cluster_a \in \mathcal{N}_{\cluster_b}~\mbox{or}~\cluster_b \in \mathcal{N}_{\cluster_a}} \mathcal{A}_{\cluster_a, \cluster_b}$;
\STATE $V^c \leftarrow \{V^c \setminus \{\cluster_a,\cluster_b\}\} \cup \{\cluster_{ab}\}$, where $\cluster_{ab} = \cluster_a \cup \cluster_b$, and $n_c = n_c - 1$;
\STATE For all $\cluster_c$, such that $\cluster_a \in \mathcal{N}_{\cluster_c}$ or $\cluster_b \in \mathcal{N}_{\cluster_c}$, add $\cluster_{ab}$ to $\mathcal{N}_{\cluster_c}$, and compute the affinity $\mathcal{A}_{\cluster_{ab}, \cluster_c}$;
\STATE Find the $K^c$-nearest clusters for $\cluster_{ab}$ in the set $\mathcal{N}_{\cluster_a} \cup \mathcal{N}_{\cluster_b}$, to form $\mathcal{N}_{\cluster_{ab}}$;
\STATE Remove $\cluster_a$ and $\cluster_b$ from the neighbor sets, and remove $\mathcal{N}_{\cluster_a}$ and $\mathcal{N}_{\cluster_b}$.
\ENDWHILE
\STATE \textbf{Output:} $V^c$.
\end{algorithmic}
\end{algorithm}

\section{Proof of Theorem \ref{thm:complexity}} \label{sec:compproof}

\begin{proof} For (a), we analyze the time complexity for each part of the GDL algorithm.

\begin{enumerate}[(1)]
\item The directed graph construction has a complexity of at most $O(K n^2)$ (naive implementation). Note that $n \gg K$, and thus we omit $K$ in the complexities hereinafter.

\item The complexity of constructing initial clusters is $O(n)$, as the number of edges in the graph $G$ is $O(K^0n)$, where $K^0$ is $1$ or $2$.

\item In our clustering algorithm, we use an $n_c \times n_c$ table to store the affinities between clusters. As the initial clusters are of small sizes, we can assume $O(1)$ complexity for computing the affinity between each pair of two clusters. So, it requires a complexity of $O(n_0^2)$ to initialize the table, where $n_0$ is the number of initial clusters ($n_0 < n$).


\item In each iteration, it costs $O(n_c^2)$ to find the maximum value in the cluster affinity table. To update the cluster affinity table after merging the two clusters with maximum affinity value, we need to compute $(n_c-1)$ affinities, and each affinity is computed with complexity of $O(|\cluster_a| + |\cluster_b|)$ using Eq. (\ref{eqn:aff}) (because $\mathbf{W}$ is a sparse matrix with $K$ nonzero elements in each row). Therefore, the complexity for each iteration is at most $O(n_c n)$.




\item The number of iterations is $(n_0 - n_T)$.
\end{enumerate}

By replacing $n_0$ and $n_c$ with their upper bound $n$, a loose upper bound of the time complexity of the GDL algorithm is $O(n^3)$.

~

For (b), we can reduce the complexity in each iteration from $O(n_c n)$ in (a) to $O(n)$. We can maintain a table to store the nearest cluster of each cluster.\footnote{We can use a heap to achieve better efficiency for this part. But it is not the bottleneck for both the complexity analysis and run-time of GDL-U.} In each iteration, finding the maximum value and updating the table cost approximately $O(n_c)$. For the affinity table, the updating scheme of $\mathcal{A}_{\mathcal{C}_{ab} \rightarrow \mathcal{C}_c}$ as in Eq. \ref{eqn:lw} costs $O(n_c)$ for all the new affinities. To compute $\mathcal{A}_{\mathcal{C}_c \rightarrow \mathcal{C}_{ab}}$, the total complexity for all the new affinities is less than the complexity of computing $\mathbf{W}_{\mathcal{C}_{ab}, *} \mathbf{W}_{*, \mathcal{C}_{ab}}$, which is $O(n K)$, as $\affm_{\mathcal{C}_{ab}, *}$ is $K$-sparse in each row. $\affm_{\mathcal{C}_{ab}, *}$ is the submatrix of $\affm$ whose row indices correspond to the vertices in $\cluster_{ab}$ and column indices are from $1$ to $n$.

Finally, the total complexity for GDL-U is $O(n^2)$.

~

For (c), there are several differences in the AGDL:
\begin{itemize}
\item In (3), we use the neighbor sets of clusters instead of the cluster affinity table. The construction of all the neighbor sets costs $O(K^c n_0^2)$.

\item In (4), we need to find the maximum affinity value in the neighbor sets (with complexity of $O(K^c n_c)$ and compute $O(K^c (1 + \tau))$ affinities to update the neighbor sets with complexity of $O(K^c n)$). Because the size of the union of neighbor sets of $\cluster_a$ and $\cluster_b$ is less than $2 K^c)$, and for real data, we can assume that the number of clusters whose neighbor set includes $\cluster_a$ or $\cluster_b$ is less than $2 \tau K^c$, where $\tau$ is usually a small constant close to $1$. Therefore, the complexity for each iteration is at most $O(n)$.
\end{itemize}

So, the time complexity of the AGDL algorithm is $O(n^2)$.
\end{proof}

\section{Quantitative Results in Clustering Error for Image Clustering} \label{sec:ce}
The quantitative results, measured in CE \cite{wu2007local}, are given in Table \ref{tab:imgce}. The CE is defined as the minimum overall error rate among all possible permutation mappings between true class labels and clusters. A smaller CE value indicates a better clustering result.

\begin{table}[tpb]
\center
\caption{Quantitative clustering results in CE on real imagery data. A smaller CE value indicates a better clustering result. The results shown in a boldface are significantly better than the others, with a significance level of 0.01.} \label{tab:imgce}
{
\begin{tabular}{|l||c|c|c|c|c|c|c||c|c|c|}
\hline {Dataset} & k-med & Link  & G-Link & NCuts & NJW-SC &  DGSC & STSC & Zell & GDL-U & AGDL \\
\hline
COIL-20          & 0.401 & 0.677 &  0.213 & 0.246 &  0.228 & 0.201 & 0.158 & 0.187 & \textbf{0.142} & \textbf{0.142} \\
COIL-100         & 0.570 & 0.819 &  0.394 & 0.462 &  0.411 & 0.396 & 0.391 & 0.351 & \textbf{0.267} & \textbf{0.269} \\
{USPS}           & 0.607 & 0.874 &  0.252 & 0.459 &  0.354 & 0.255 & 0.421 & 0.332 & \textbf{0.246} & \textbf{0.246} \\
{MNIST}          & 0.577 & 0.776 &  0.162 & 0.405 &  0.432 & 0.230 & 0.305 & 0.400 & \textbf{0.150} & \textbf{0.150} \\
Yale-B           & 0.728 & 0.847 &  0.376 & 0.273 &  0.270 & 0.237 & 0.205 & 0.464 & \textbf{0.197} & \textbf{0.197} \\
FRGC             & 0.728 & 0.753 &  0.664 & 0.565 &  0.596 & 0.595 & 0.580 & 0.560 & \textbf{0.548} & \textbf{0.551} \\
\hline
\end{tabular}
}
\end{table}

\section{Outlier Elimination for Object Matching} \label{sec:olagdl} 

\begin{figure}[t]
\center
\includegraphics[width=0.366\columnwidth]{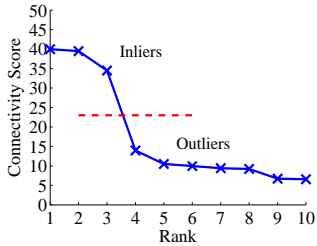}
\caption{The connectivity scores of clusters sorted in descending order. The threshold for separating inliers and outliers is shown in a red dash line.} \label{fig:olcurve}
\end{figure}

For AGDL, we observe that there are many inedges and outedges inside a cluster of inliers, while less edges inside a cluster of outliers because outliers are in low density regions. Inspired by this, we define the connectivity score of a cluster $\cluster$ as\\ $\sum_{i \in \cluster} \left[\deg^-_i (\cluster) + \deg^+_i (\cluster)\right]$. We find that there are always large differences between the scores of inlier clusters and outlier clusters (see Fig. \ref{fig:olcurve}). Therefore, we rank the final clusters by their connectivity scores. Namely, we sort their scores in descending order, and then search the largest gap between two consecutive scores. The set of clusters are divided into two subsets without intersection. The subset of clusters with small scores is treated as the collection of outliers and removed. For ACC and GS, we use their default methods for outlier elimination.

%

\end{document}